\documentclass[11pt]{article}

\usepackage[T1]{fontenc}
\usepackage[utf8]{inputenc}
\usepackage{mathptmx}
\usepackage{courier}
\usepackage[margin=1.05in]{geometry}
\usepackage{amsmath,amssymb,amsthm}
\usepackage{xcolor}
\usepackage[numbers,sort&compress]{natbib}
\usepackage[colorlinks=true,linkcolor=blue!50!black,citecolor=blue!50!black,urlcolor=blue!50!black]{hyperref}
\usepackage{graphicx}
\usepackage{booktabs}
\usepackage{caption}
\usepackage{enumitem}
\usepackage{placeins}
\usepackage{tikz}
\usetikzlibrary{matrix,positioning,arrows.meta}
\usepackage{array}

\newtheorem{criterion}{Criterion}

\setlist[itemize]{itemsep=2pt, topsep=3pt, leftmargin=1.3em}
\setlist[enumerate]{itemsep=2pt, topsep=3pt, leftmargin=1.6em}

\title{\textbf{Exactness at Inference:\\ A Representational Criterion for
Out-of-Distribution Generalization}}

\author{
Filipe Marinho Rocha\thanks{Corresponding author:
\texttt{filipemarocha@gmail.com}.}\;$^{1,2}$
\and In\^{e}s Dutra$^{1,3}$
\and V\'{i}tor Santos Costa$^{1,2}$
\and Lu\'{i}s Paulo Reis$^{4,5}$
}

\date{
\small
$^{1}$Faculdade de Ci\^{e}ncias da Universidade do Porto, Porto, Portugal\\
$^{2}$INESC TEC, Campus da FEUP, Porto, Portugal\\
$^{3}$CINTESIS@RISE-Health, Porto, Portugal\\
$^{4}$Faculdade de Engenharia da Universidade do Porto,
Rua Dr.\ Roberto Frias, Porto, Portugal\\
$^{5}$LIACC, FEUP, Porto, Portugal
}

\hypersetup{pdftitle={Exactness at Inference: A Representational Criterion for Out-of-Distribution Generalization},
  pdfauthor={Filipe Marinho Rocha, In\^{e}s Dutra, V\'{i}tor Santos Costa, Lu\'{i}s Paulo Reis}}
\begin{document}
\maketitle

\begin{abstract}
\noindent
A model generalizes outside its training distribution only when it computes a
representation structurally equivalent to the underlying data-generating
mechanism, rather than a fitted approximation calibrated on data. Structural
equivalence is necessary for exactness both in-distribution and
out-of-distribution, and extrapolation is governed by this \emph{exactness at
inference}, independent of how the representation is realized,
provided the structural equivalence holds.

Tensor Logic exemplifies this independence. A zero-temperature tensor
contraction achieves strict mathematical equivalence to discrete logic,
executing exact deductions in place, in the tensor language itself, with no artefact extracted. At zero temperature the inferred tensors are Boolean and the
embeddings are orthonormal, so what remains continuous is the arithmetic rather
than the values. The realization still has strict geometric limits. Lacking
infinite functional recursion, it is equivalent only to Datalog and not to full
Prolog, and while it is exact over closed domains, binding variables to novel entities requires external-memory indirection or explicit rule
extraction, since a fixed tensor cannot dynamically expand its index.

Evaluated broadly, this criterion of exactness is sharper than it appears in
three ways. First, it does not require a discrete representation: a sum-product
tensor contraction computing an exact marginal satisfies it across $[0,1]$,
whereas a fitted Neural Network thresholded to output a hard label does not. Second, it
does not require an extracted symbolic expression: a continuous relaxation
collapsed in place works natively as a rule. Third, it applies strictly to
inference time, which allows approximate continuous search during training.

Under this criterion, Logic Tensor Networks fail, whereas differentiable
ILP and zero-temperature Tensor Logic pass. The two long-known failure modes of
deep networks are the continuous and relational expressions of a shortfall in
exact representability: piecewise-affine extrapolation divergence and an inability to bind novel entities. For hybrid architectures we introduce an
inference-path propagation rule: the output inherits the representational bounds
of every fitted estimator along its path. This explains, for example,
which coordinate axes fail in equivariant models, and predicts the empirical
divide on ARC-AGI, where inductive programs dominate precise compositions and
transductive networks dominate perceptual matching. An epistemic corollary
follows: only models that commit to a hypothesis within an exact class can
certify what their training data leaves underdetermined. We give an example: on a law-derived evaluation partition, an exact model identifies the $56.3\%$ of distant queries that are definitively
answerable, whereas deep network ensembles exhibit false confidence and distance-based
uncertainty metrics rank them backwards.

These findings reveal an upper bound on current architectures: the common
inductive biases, from hardcoded symmetries to external memory, achieve
exactness only because humans manually inject representations structurally
equivalent to the generating processes. Overcoming this limitation requires
architectures capable of inducing exact representations autonomously, rather
than fitting continuous surrogates constrained to representations such as the
piecewise-affine maps of Multilayer Perceptrons (MLPs), which compose Neural Networks and are incorrect unless the generating mechanism is itself of that form, and it seldom is.
While such surrogates can approximate in-distribution data, their residual errors, even when driven to the arithmetic floor on the training data, diverge out-of-distribution and compound under composition.
\end{abstract}

\section{Introduction}
\label{sec:intro}

A model generalizes outside its training distribution only when it computes a
representation structurally equivalent to the underlying data-generating
mechanism, rather than a fitted approximation calibrated on data. That is the
claim this paper defends, and two failures of deep networks, documented for long
enough that neither is news, are where it is most visible. A Neural Network built from
affine
layers and piecewise-linear activations extends the affine map attached to its
outermost region indefinitely, so it cannot continue a target that curves
\citep{hein2019relu, xu2020how}. And a Neural Network that represents an entity by a
free parameter fitted to that entity has nothing to say about an entity it has
never seen, so it cannot apply a universally quantified rule to it
\citep{marcus1998rethinking, marcus2001algebraic, teru2020grail}.

The usual readings of these results are that Neural Networks need more data, more
scale, or a better inductive bias. We think the more useful reading is
representational and that it applies to both at once: what a model can compute
correctly outside its training data is what it can compute \emph{exactly}, and
everything
else is an approximation whose validity was established on the training
distribution and does not transfer. Exactness is in turn a representational
condition rather than a numerical one: an operation is exact when the
representation it executes is structurally equivalent to the mechanism that
generated the data, and a surrogate calibrated on a region is not rendered exact
by being accurate on that region, which is why the condition binds inside the
training distribution as well as outside it. This is not a claim about accuracy.
A model can be highly accurate and still fail the criterion, and the difference
shows up precisely where accuracy was never measured, on Out-of-Distribution
(OOD) data.

This paper has two purposes: it brings together the account of the two representational limits in a form we could not find presented anywhere, and it
states a criterion we believe to be new, and uses it to do work that the
existing vocabulary cannot. Sections marked as such are
review, and \S\ref{sec:established} says plainly what belongs to whom.

Our own contribution is the criterion, and its consequences.

\begin{enumerate}[label=(\roman*)]
\item \textbf{Exactness, not discreteness} (\S\ref{sec:criterion}). What must
hold at inference is that the operation executes a representation
\emph{structurally equivalent} to the data-generating mechanism, so that it
computes the intended relation rather than approximating it in a region where
there is training data to calibrate on. Structural equivalence is necessary for
exactness in both regimes and not only beyond the training data, since a
surrogate calibrated on a region
carries a residual there too and is merely tolerable where it was fitted
(\S\ref{sec:datahungry}). Discreteness is what exactness reduces to when the
intended semantics is classical logic; it is not the criterion itself. A sum-product tensor contraction computing an exact marginal is
continuous-valued and exact; a fitted Neural Network thresholded to a hard label
is discrete-valued and not exact.

\item \textbf{A sorting that cuts across the usual taxonomy}
(\S\ref{sec:sorting}). Applied to neuro-symbolic systems the criterion does not
separate them by whether they contain a logical component. Logic Tensor Networks (LTN) \citep{badreddine2022logictensor} fail it, $\partial$ILP \citep{evans2018learning}, a differentiable form of Inductive Logic Programming (ILP), passes it by extracting a program,
and Tensor Logic \citep{domingos2025tensorlogic} at zero temperature passes it
over a closed domain without extracting anything,
since its equations are already equivalent to rules, though an open domain
returns it to one of the other two routes (\S\ref{sec:sorting}). What matters is
not whether a rule is extracted but whether an approximation using a wrong
representation is consulted when the inference is made.

\item \textbf{A propagation rule for composed architectures}
(\S\ref{sec:composition}). Nearly every Deep Learning (DL) architecture in use
contains a Multilayer Perceptron (MLP) somewhere, and the conclusion that all of
them inherit its limits is false. A quantity inherits the limits of every fitted
component on \emph{its} inference path, and not of components outside this path.
In the case of an equivariant model, this predicts in which axis the model
fails, why external-memory indirection buys binding, and why energy conservation and trajectory accuracy come
apart in the same Neural Network, and it is answerable by inspecting an architecture
before running anything.

\item \textbf{One shortfall, two faces} (\S\ref{sec:continuous-face},
\S\ref{sec:relational-face}). The continuous and relational failures are the same
shortfall in \emph{exact representability}, the class of functions and relations
an architecture computes exactly rather than approximates, and they appear in the
continuous and discrete domains respectively. Both failures are established results and the
unification is the novel contribution; we argue it earns its place by predicting
where each failure does \emph{not} occur, and by scoping each against its known
counterexamples.

\item \textbf{The epistemic corollary} (\S\ref{sec:epistemic}). A model can know
that it is right only where it is using the correct underlying rule, since
abstention is a statement about what the training data determines and
determination is relative to a hypothesis class. On a radial chirp, recovering the
exact rule fixes $56.3\%$ of the distant queries from the training data alone,
while deep network ensembles fail confidently on those same queries and
distance-based estimators flag them as the most uncertain, so an estimator added
on top abstains exactly where the answer is already determined. Only a system
that commits to an exact hypothesis carries a reliable internal signal of its own
recovery.

\item \textbf{Epistemic dependence} (\S\ref{sec:dependence}). Every known repair
supplies exactness by having a person program the structure into the architecture
in advance: the group in an equivariant architecture, the indirection in an
external-memory controller, the symplectic form in a Hamiltonian Neural Network. Each
works only where the designer supplied the right structure. If
a model does not induce the structure, someone must inject it, and that is a
bound on what scaling alone reaches. We also argue that the epistemic dependence
commonly charged against symbolic systems is universal rather than specific to
them, and that what distinguishes the paradigms is whether the dependence is
visible and measurable.
\end{enumerate}

Our representational criterion should not be confused with the \emph{representation bottleneck} of \citet{deng2022representation},
which names the finding that Neural Networks encode interactions of low and high
complexity but not intermediate ones, a claim about interaction order and not
about extrapolation. Nor should it be confused with the \emph{relational
bottleneck} of \citet{webb2024relational}, which is a beneficial architectural
constraint restricting information flow to relations, close to the opposite of a
limitation.

\section{What Is Already Established}
\label{sec:established}

Before introducing a new representational criterion for generalization, it is
necessary to delineate the boundaries of what the field has already established.
This section is strictly a review of prior work. Its purpose is to make the
remainder of our argument checkable, since the value of any proposed criterion
depends heavily on its being rigorously distinguishable from the several
phenomena it resembles and seeks to explain.

Currently, the literature recognizes two primary modes of OOD failure in DL
architectures. In the continuous domain, models fail to extrapolate because they
are trapped by the asymptotic geometry of piecewise-affine maps. In the
relational domain, models fail at open-domain logic due to an inability to
dynamically bind variables to novel entities. Historically, these have been
treated as separate pathologies, diagnosed by different communities using
different vocabularies. By assembling the accepted scope, mechanisms and known
counterexamples of both limits into a single account, we separate
the empirical symptoms, which are widely documented, from our theoretical
diagnosis. Laying this foundation ensures that when we introduce our criterion of
\emph{exactness at inference}, it is clear exactly which previous theoretical gaps it
bridges, why simply scaling data or parameters does not resolve these limits, and
why a unified representational theory is required.

\subsection{The continuous limit}

This limit is established in the literature. \citet{hein2019relu} show that ReLU
networks produce
arbitrarily high-confidence predictions far from the data, because the outermost
regions of the partition are unbounded and the affine map on each is prolonged
without termination. \citet{xu2020how} prove the sharper statement that ReLU MLPs
converge to linear functions along any direction from the origin, so they do not
extrapolate most nonlinear targets, but that they do learn linear targets
given a sufficiently diverse training distribution. That success on linear
targets is not a coincidence of optimization. They learn such targets because the
piecewise-affine representation native to an MLP, formalized in
\S\ref{sec:pwa}, is structurally equivalent to a linear generating mechanism.
This condition is important, and we return to it as the positive control of
\S\ref{sec:boundary}. The partition itself is characterized by
\citet{montufar2014number}.

Two remarks on the scope of these results are needed, because both are
often overstated. First, \citet{xu2020how} argue via the neural
tangent kernel, which describes what gradient descent converges to, whereas the
polytope picture describes what the architecture can represent. The two agree
here, but they are different claims, and only the second is what our criterion
needs. Second, the theorem is about ReLU. Extending it to the smooth activations
used in practice is not automatic, and \S\ref{sec:activations} reports a case
where the intuition drawn from a single unit gives the wrong answer for the
Neural Network: a GELU unit is asymptotically linear on one tail and asymptotically zero
on the other, which suggests that a GELU network should diverge on one side and
flatten on the other, and the measurement shows that it diverges on both
(Figure~\ref{fig:asymptotics}).

\subsection{The relational limit}

Also established in the literature, and older.
\citet{mccarthy1988epistemological} named the phenomenon
\emph{propositional fixation}. \citet{fodor1988connectionism} gave the
systematicity argument that the modern versions instantiate.
\citet{marcus1998rethinking} showed that Neural Networks whose representation of an item
is a free parameter cannot extend a universally quantified mapping to items
outside the training space, and \citet{marcus2001algebraic} develops the argument
at book length. \citet{teru2020grail} state the modern embedding version plainly:
methods that learn latent representations of entities ``do not explicitly capture
the compositional logical rules'' and ``are limited to the transductive setting,
where the full set of entities must be known during training.''
\citet{greff2020binding} give the general form as a binding problem.
\citet{barcelo2020logical} supply the precise expressiveness result for GNNs,
which is that they capture a fragment of First-Order Logic (FOL) with two
variables and counting, given a readout, more than propositional and less than
first-order.

Two constructive results bound the claim from the other side.
\citet{smolensky1990tensor} showed that connectionist variable binding is
constructible in principle via tensor products, and \citet{webb2021esbn} built a
Neural Network that binds and generalizes to novel entities. \S\ref{sec:right-scope}
takes both seriously.

For Large Language Models (LLMs) the pattern is documented empirically.
\citet{mirzadeh2024gsmsymbolic} conclude that models ``cannot perform genuine
logical reasoning'' and instead ``replicate reasoning steps from their training
data,'' and \citet{dziri2023faith} report the corresponding compositional
collapse.

\subsection{And what is not established}

Three things, all of which our criterion supplies.

The literature does not offer a criterion for when a hybrid system inherits the
limitation. That a system contains a logical component says nothing; the
sorting in \S\ref{sec:sorting} shows that systems containing a full first-order
syntax fail while systems containing no separately extracted artefact pass. The
nearest antecedent is \citet{ahmed2022semantic}, who observe that loss-based
neuro-symbolic methods ``cannot guarantee that the predictions will be consistent
at test time'' and construct a layer that guarantees it instead. That is our
distinction, made for constraint satisfaction, and we generalize it.

The literature does not distinguish \emph{exactness from discreteness}. This is
the crux, and getting it wrong in either direction misclassifies real systems, as
\S\ref{sec:not-discreteness} shows.

\citet{deletang2023chomsky} come close
from the empirical side, showing across 20,910 models that architecture class
forecasts OOD generalization and that ``even extensive amounts of
data and training time never lead to any non-trivial generalization, despite
models having sufficient capacity''. \citet{kim2025standard} comes close from the
analytic side, deriving both the continuous approximation limit and the difficulty
of universal quantification over unbounded domains from the assumptions of the
Universal Approximation Theorem (UAT). We regard our contribution as the criterion, not
just the observation that the two failures meet.

\section{The Continuous Face}
\label{sec:continuous-face}

\subsection{Neural Networks as piecewise-affine partitions}
\label{sec:pwa}

A feedforward Neural Network consists of affine transformations interspersed with
nonlinear activations. When those activations are piecewise linear, or when
smooth activations are analysed at their local linear limits, the Neural Network as a
whole, regardless of depth or width, partitions the input space into a large
number of convex regions \citep{montufar2014number}. Within a single region
$\Omega_i$ the Neural Network applies one affine map,
\begin{equation}
f(x) = W_i x + b_i \quad \text{for} \quad x \in \Omega_i ,
\end{equation}
with $W_i$ and $b_i$ the effective weight matrix and bias for that region. Every
claim in this section follows from that fact and from where the regions are
placed.

\subsection{Inside the data: what more data buys, and what it does not}
\label{sec:datahungry}

Within the training domain the Neural Network minimizes empirical risk by packing
linear regions densely where the target bends. This tightly woven polygonal chain
makes the approximation look smooth and locally accurate.

\begin{figure}[htbp]
\centering
\includegraphics[width=\textwidth]{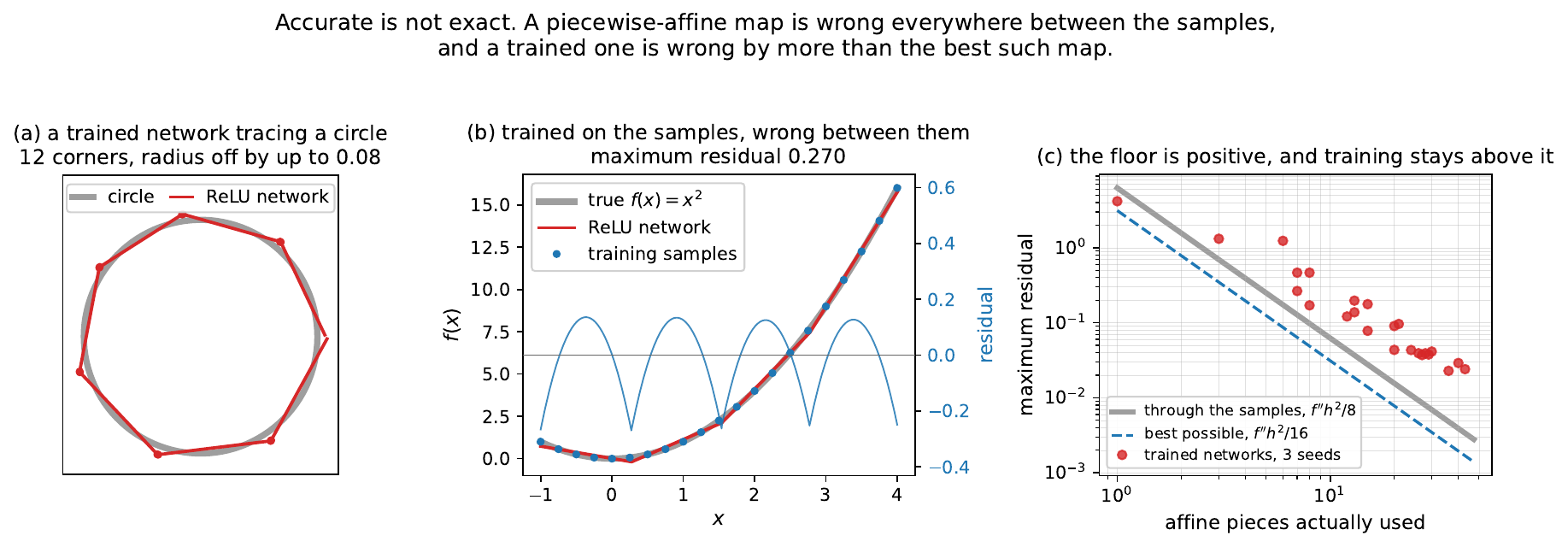}
\caption{Accurate is not exact, measured. (a) A ReLU network with one hidden
layer of twelve units, trained to trace the unit circle from the angle of a
point on it: the output is a polygon of twelve corners whose radius is wrong by
up to $0.08$, and whose corners do not lie on the circle, the map being fitted
rather than inscribed. (b) A Neural Network of ten units trained on samples of
$f(x)=x^{2}$ over $[-1,4]$; the residual, on the right axis, is zero only at
isolated crossings and curved everywhere between them, reaching $0.270$.
(c) Maximum residual against the number of affine pieces the trained Neural Network
actually uses, three seeds at each of eight widths, against two analytic curves:
the error of the piecewise-affine map through the samples, $f''h^{2}/8$, and the
error of the best piecewise-affine map of the same number of pieces,
$f''h^{2}/16$, which is the chord bound halved by the Chebyshev equioscillation
theorem. Every trained Neural Network lies above the second, which is the floor
its architecture imposes, and all but one above the first. Refining the partition
lowers the floor as $h^{2}$ and never to zero.}
\label{fig:residual}
\end{figure}

Accurate is not exact, and we state the gap precisely, because the
in-distribution part of the criterion lives in it. Between two adjacent
breakpoints the Neural Network is affine while the target curves, so on an interval of
width $h$ the residual against a twice-differentiable target is of order
$h^{2}\max|f''|/8$ and vanishes only where $f''=0$. Interpolating the training
points exactly does not remove it. It pins the chain near the sample points and
leaves the chords between them, in the way that a polygon meets a circle at a few
points and misses it everywhere else, and no number of corners makes the polygon
a circle. Figure~\ref{fig:residual} measures this on trained Neural Networks. A Neural Network
given the angle of a point on the unit circle returns a twelve-cornered polygon
whose radius is wrong by up to $0.08$, and a Neural Network trained on samples of
$x^{2}$ carries a residual that is zero only at isolated crossings and curved
between them. The same figure separates the floor from what training reaches: for
a map of $n$ equal pieces the residual cannot fall below $f''h^{2}/16$, and a map
through every sample gives $f''h^{2}/8$, twice that. The factor of two is the
Chebyshev equioscillation theorem in its simplest instance, since the best affine
approximant on a subinterval is the chord displaced by half its maximum
deviation (\S\ref{sec:derivations}). The trained Neural Networks lie above the
first at every width, by between $1.3$ and $14.9$ times, and above the second in
twenty-three of twenty-four runs. The architecture sets a floor and
the optimizer does not reach it. The residual is therefore non-vanishing on
the whole continuous support rather than only outside the data, and what a dense
training distribution buys is that it stays small enough to disappear into the
aggregate error. Two later arguments rest on this. The same residual is what
diverges once the input leaves the region where the spacing was made small, the
subject of the next two subsections, and what compounds when fitted modules are
composed (\S\ref{sec:composition}).

This geometry is a sufficient account of why such models are data-hungry, and it
needs no appeal to optimization or to scale. Lacking any deductive mechanism, the
Neural Network can raise fidelity only by placing boundaries where the target actually
curves. In high dimensions the volume to be covered grows exponentially, so the
number of examples required to position those boundaries grows with it, which is
the \emph{curse of dimensionality} \citep{bellman1957dynamic} in its approximation-theoretic
form. The count follows from the bound above. Driving the residual below
$\varepsilon$ requires a cell of width of order $\sqrt{\varepsilon}$ along every axis,
so covering a $d$-dimensional domain takes on the order of $\varepsilon^{-d/2}$ affine
pieces, and matching upper and lower complexity bounds of this form are established
for ReLU networks by \citet{yarotsky2017error}. Where
data is sparse the regions stay large and rigid, drawing straight lines across
curved space.

\begin{figure}[htbp]
\centering
{\small A ReLU network cannot bend. It can only place breakpoints (vertical lines) and run
straight between them, and it needs data to know where to put them.\par}\medskip
\includegraphics[height=2in]{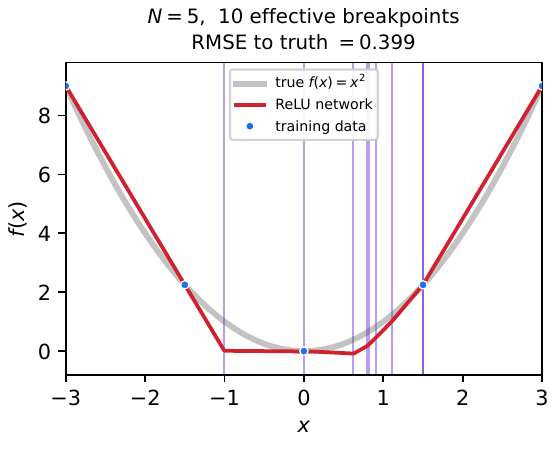}\hspace{0.6em}%
\includegraphics[height=2in]{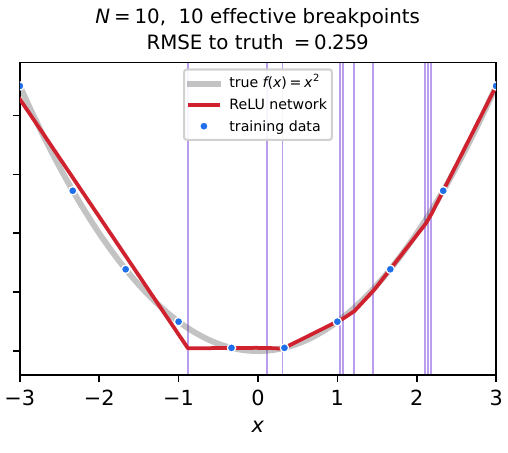}\\[0.4em]
\includegraphics[height=2in]{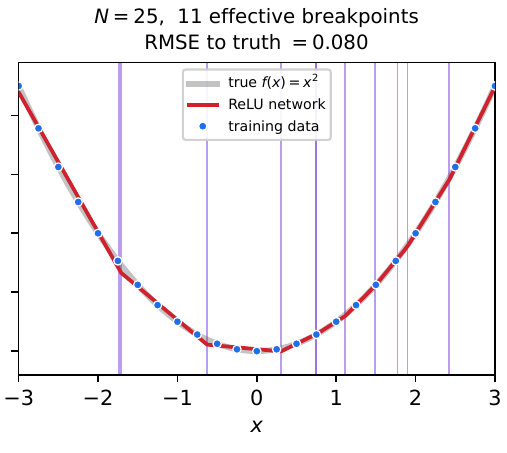}\hspace{0.6em}%
\includegraphics[height=2in]{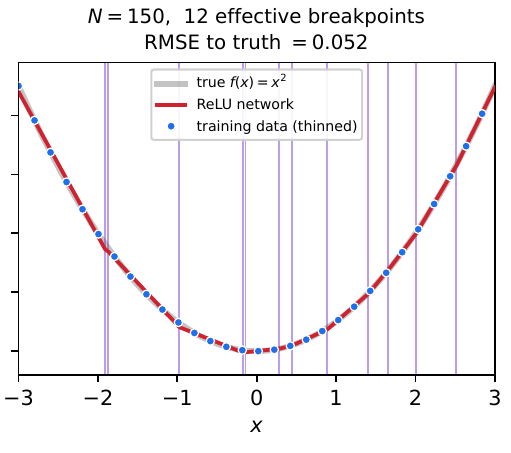}
\caption{What more data buys inside the training region. One-hidden-layer ReLU
networks of fixed width fit $f(x)=x^{2}$ on $[-3,3]$ from $N$ evenly spaced
points; vertical lines mark the breakpoints the trained Neural Network actually uses.
The count barely changes across the panels while the error falls by almost an
order of magnitude, because the data is not buying additional breakpoints, which
the architecture fixes, but better placement of the ones it already has.}
\label{fig:datahungry}
\end{figure}

It also separates two things that are easily conflated, capacity and placement,
and the separation is a controlled dissociation. Figure~\ref{fig:datahungry} fits
$f(x) = x^2$ with one-hidden-layer ReLU networks of \emph{fixed} width from an
increasing number of evenly spaced points, and marks the breakpoints the trained
Neural Network actually uses, meaning those whose slope change exceeds $0.05$ against a
true slope ranging over $[-6,6]$. Across $N = 5, 10, 25, 150$ the usable count
runs $10, 10, 11, 12$ while the error falls from $0.399$ to $0.052$, a factor of
about eight. The number available is fixed by the architecture at twelve, since
each hidden unit contributes at most one, and it does not grow when more data
arrives. What data buys is knowing where to put them. The rightmost panel is at
the architectural ceiling, so the eightfold improvement over the leftmost cannot
have been bought with additional pieces; two extra usable breakpoints accompany
it, and the rest is placement.

This matters twice for what follows. It is the mechanism behind the
\emph{observability} condition of \S\ref{sec:boundary}: data is what pins down
which member of the representable class the Neural Network has settled on, and it does so
by positioning pieces rather than by supplying new ones. And it is a small,
controlled instance of the scaling argument of \S\ref{sec:dependence}: enlarging
the fitted part of a system changes the quality of an approximation and leaves the
class being approximated within exactly where it was.

\FloatBarrier

\subsection{Extrapolation as asymptotic extension}
\label{sec:extrapolation}

The structural vulnerability of the partition appears at the boundary of the
training manifold. There are no data points outside that domain to prompt the
formation of further bounding hyperplanes, so the outermost regions
$\Omega_{\text{ext}}$ extend outward indefinitely \citep{hein2019relu}. An
OOD input falls into one of these unbounded regions, and the
Neural Network applies the terminal state that defined the boundary of the data,
whether an affine projection or a saturated constant, projecting it endlessly
into space the data never constrained. This is the mechanism behind confident and
badly wrong predictions off-distribution \citep{xu2020how}, and we call it
\emph{piecewise-affine extrapolation divergence}, both to name the continuous
face of the shortfall for the propagation rule of \S\ref{sec:composition} and to
separate it from the regime of \S\ref{sec:scope}, where a bounded representation
converges toward a constant instead.

We state the point in the form that matters for this paper. The Neural Network
does not fail to answer. It answers with the same fluency it shows on familiar
input, because the operation it performs off-distribution is the same operation
it performs on it.

\subsection{Case study: the quadratic, across three activations}
\label{sec:activations}

\begin{figure}[htbp]
\centering
\includegraphics[width=\textwidth]{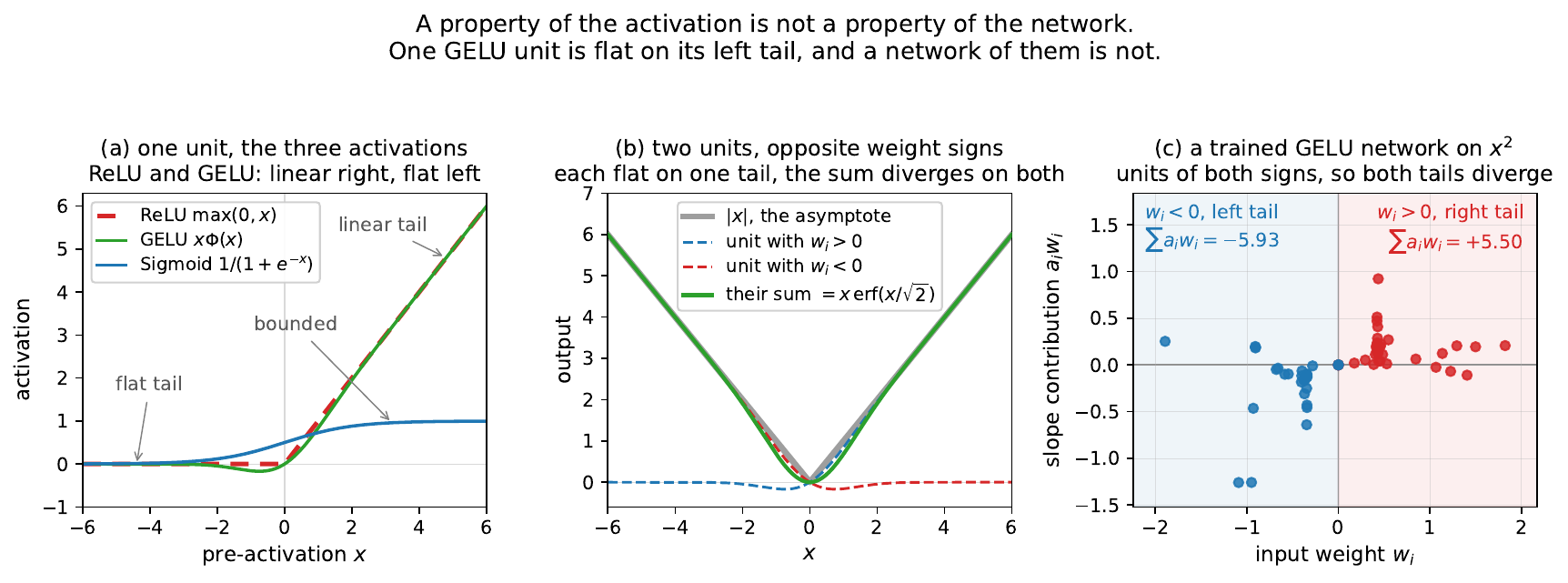}
\caption{The three activations at unit level, and the step that the unit-level
reading does not survive. (a) ReLU and GELU are asymptotically linear on one tail
and flat on the other, while Sigmoid is bounded on both. (b) Two GELU units whose
input weights carry opposite signs are each flat on one tail, and their sum is
$x\,\mathrm{erf}(x/\sqrt{2})$, with $\mathrm{erf}$ the error function, which grows
like $|x|$ on both (\S\ref{sec:derivations}). (c) A
one-hidden-layer GELU network of $64$ units trained on $f(x) = x^2$ over $[-3,3]$
by full-batch Adam, with each hidden unit's asymptotic slope contribution
$a_i w_i$ plotted against its input weight $w_i$. Units of both signs are
present, the two groups sum to the left and right tail slopes of the Neural Network, and
those sums agree with the slopes measured directly at $x = \pm 200$. Panels (a)
and (b) are exact evaluations of closed-form functions, panel (c) is a trained
Neural Network.}
\label{fig:asymptotics}
\end{figure}

\begin{figure}[htbp]
\centering
\includegraphics[width=\textwidth]{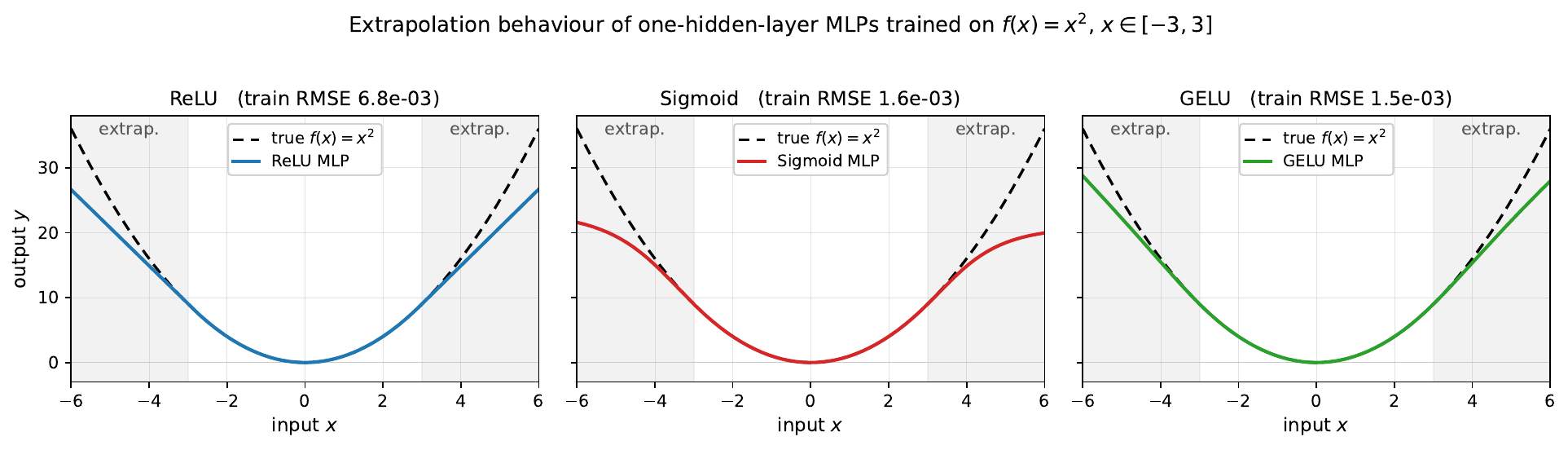}
\caption{Three one-hidden-layer MLPs (64 units, full-batch Adam) fitting
$f(x)=x^2$ on $[-3,3]$ and evaluated on $[-6,6]$; shaded bands mark the
extrapolation region. All three fit the training domain almost exactly and all
three depart from the true function immediately outside it, in a manner set by
the asymptotic behaviour of the activation rather than by the shape of $f$. GELU
diverges on both tails, for the network-level reason given in the text.}
\label{fig:activation}
\end{figure}

Consider $f(x) = x^2$ on $x \in [-3,3]$, evaluated on $[-6,6]$. Because the neural
network resolves into a continuous piecewise-linear function, its regions are
defined by first-degree polynomials, so an MLP has no intrinsic mechanism
for quadratic growth, for higher-degree polynomials, or for transcendental
functions such as $\sin(\cdot)$ and $\cos(\cdot)$. This is a representational
limitation of standard Neural Networks, and it holds before any question of
optimization or scale arises. It can join short affine segments into a
convincing local approximation inside a dense data manifold, and outside it the
approximation is governed entirely by the asymptotic behaviour of the activation
rather than by the shape of $f$. Figure~\ref{fig:asymptotics} shows the three
activations at the level of a single unit, and Figure~\ref{fig:activation} shows
three trained Neural Networks. The three activations fail differently, and the
difference is set by what each unit does far from the origin.

\textbf{ReLU} extends the terminal slope, since a unit is exactly affine beyond
its own breakpoint (Figure~\ref{fig:asymptotics}a). If the tangent near $x = 3$ has slope
about $6$, the prediction beyond it is approximately $6x - 9$, growing linearly
where the true function accelerates, so the error grows without bound.

\textbf{Sigmoid} fails differently, and more gently, because the unit is bounded
on both tails (Figure~\ref{fig:asymptotics}a). Pre-activations saturate one
by one as the input leaves the training range, and the output bends toward a
plateau instead of continuing to accelerate, so the error is bounded rather than
unbounded. The plateau does not sit at the boundary value $f(\pm 3) = 9$; it is
set by the sum, over the hidden units still active at that extreme, of their
output-layer contributions. In the Neural Network measured here the two tails level off
near $20$ and $23$ rather than at $9$, and they need not agree with each other,
since a different subset of units survives at each extreme.

\textbf{GELU} is the interesting case, and the one where the natural inference is
wrong. Modern Transformers rely on $x\Phi(x)$, with $\Phi$ the standard normal
CDF. A single GELU unit is asymptotically linear for large positive
pre-activations and asymptotically zero for large negative ones, and it could be
tempting to conclude that a GELU network should diverge on one side and
flatten on the other. It does not: it diverges on both. For a hidden unit
with input weight $w_i > 0$ the pre-activation grows large and positive as
$x \to +\infty$, contributing a linear term, and large and negative as
$x \to -\infty$, contributing nothing; for $w_i < 0$ the roles are reversed.
Fitting an \emph{even} target such as $x^2$ generically requires hidden units of
both weight signs, so each tail of the extrapolation region is dominated by a
different subset of units, and both subsets contribute an unbounded linear term
to their own tail. The elementary case is two units of opposite input-weight
sign, whose sum is $x\,\mathrm{erf}(x/\sqrt{2})$, with $\mathrm{erf}$ the error
function (\S\ref{sec:derivations}). It grows like $|x|$ on both sides, since
$\mathrm{erf}(z) \to \pm 1$ as $z \to \pm\infty$, even though neither unit grows
on more than one (Figure~\ref{fig:asymptotics}b), and a trained Neural Network is
assembled the same way.
Of the $64$ hidden units fitting $x^2$ on $[-3,3]$ in
Figure~\ref{fig:asymptotics}c, $34$ carry a positive input weight and $30$ carry
a negative one, and the two groups sum to tail slopes of $+5.50$ and $-5.93$, so
the Neural Network grows without bound in both directions. Evaluated at $x = 200$ and
$x = -200$ it returns $1.10 \times 10^{3}$ and $1.18 \times 10^{3}$, where the
target is $4 \times 10^{4}$ at both, which is linear divergence on each tail
against quadratic growth. The word \emph{generically} is doing real work, and we
measured what it covers: across widths from $4$ to $128$ and three seeds, $14$ of
$15$ trained networks carried units of both signs and diverged on both tails, and
the single exception, the worst fit in its row, ended with every input weight
positive, so its left tail returns to the output bias rather than diverging. That
exception is the convergence regime of \S\ref{sec:scope} rather than a bounded
extrapolation, so both outcomes lie inside the account given here.

This matters twice over. GELU inherits ReLU's unbounded divergence rather than
gaining Sigmoid-like boundedness on either side, which is the worse asymptotic
failure mode and applies directly to contemporary practice, since GELU and the
closely related SiLU and Swish are standard in Transformer architectures. And it
is a reminder that properties of an activation function do not transfer directly to properties of a Neural Network built from it, which is exactly why
extending \citet{xu2020how} from ReLU to smooth activations requires the
measurement rather than the intuition. Within the window measured the GELU network
in fact tracks the true quadratic more closely than the other two, since two
opposing linear tails approximate a parabola tolerably over a bounded range; the
divergence is asymptotic and widens further out.

\FloatBarrier

\subsection{The boundary condition, which is the argument's positive control}
\label{sec:boundary}

The condition on the limit is representational. A Neural Network built
from affine layers and piecewise-linear activations represents exactly one class
exactly: the piecewise affine functions, each piece a first-degree polynomial.
When the generating mechanism lies inside that class, the Neural Network can recover it
and extrapolate indefinitely far from any training point, because the map it
prolongs beyond the outermost region is the correct continuation rather
than an artefact of where the data stopped.

The canonical case is $f(x) = |x|$, which is $x$ for $x \ge 0$ and $-x$
otherwise. It is piecewise affine with a single vertex, and a ReLU network
trained on a bounded interval containing that vertex recovers it and extrapolates
with essentially zero error. This is the same conditional \citet{xu2020how} prove
for linear targets, and it is why they report successes alongside failures in
extrapolation. A second and distinct exception
arises for bounded targets: where the true function saturates, a saturating
activation's plateau cannot diverge far from it, so the error stays small even
though nothing was learned about the region.

Two distinct conditions must therefore hold together, representability and
observability, and separating them matters because they are easily conflated.
\emph{Representability} requires that the target mechanism lie within the
hypothesis class the architecture can express. \emph{Observability} requires
the training data to determine the correct member of that class. For a
piecewise-affine network this means that every vertex of the target lies inside
the training region: the outermost affine pieces of the target are then observed
on segments of positive length, and the pieces the Neural Network prolongs
outward coincide with them. A ReLU network can represent a sawtooth with fifty
teeth exactly, but trained on three of the teeth it does not extrapolate the
remaining forty-seven, because the vertices it never observed are not in the
data.

This conditional success is what the representational argument rests on. Were the
deficit caused by something diffuse, insufficient data, imperfect optimization, or
inadequate scale, failure would be broadly uniform. It is not. Failure is
predicted precisely by whether the generating mechanism falls outside the
representable class, and success returns the moment it falls inside it. That is a
controlled dissociation, and it isolates representation as the cause rather than
merely a correlate. The cases where a Neural Network extrapolates perfectly
consequently do not contradict this account. They are its positive control.

What follows is a sharper claim than the one usually made. DL does not fail at
extrapolation as such. It fails specifically at extrapolating any generating
mechanism that is not piecewise affine, and the mechanisms we actually care about,
whether the inverse-square laws of physics, the compositional rules of abstraction
benchmarks \citep{chollet2019measure, chollet2026arc}, or the universally quantified relations of
\S\ref{sec:relational-face}, overwhelmingly are not.

\subsection{Scope: when Neural Networks converge instead of diverging}
\label{sec:scope}

The divergence picture has a real counterexample and it should be stated rather
than managed. \citet{kangocs2024extrapolate} find that Neural Networks with
high-dimensional inputs tend, as inputs become more OOD, toward a
constant that closely approximates the constant solution: the input-independent
prediction that minimizes average training loss, which \citeauthor{kangocs2024extrapolate}
call the optimal constant solution, optimal only among constants. They observe
this across eight shift datasets, on Convolutional and Transformer
architectures, under cross-entropy, squared-error and Gaussian likelihood losses.

The regimes differ, and both results are correct in their own. A softmax output is
confined to the simplex and cannot diverge; normalization layers bound
activations; and in high dimensions a sufficiently distant input drives features
toward a region where the Neural Network emits little more than its bias, which is
approximately that constant, the best input-independent prediction the training
loss admits. An unnormalized regression head on a low-dimensional input has none
of these boundaries, and diverges. We therefore scope the
divergence claim to unbounded-output regression networks without normalization,
which is the setting of Figure~\ref{fig:activation}.

What matters for the criterion is that convergence is not a different kind of event.
A Neural Network falling back to the training mean is not computing the generating
mechanism either; it has substituted a data-derived default for a law. Both
behaviours are the same shortfall in exact representability, differing in what the
architecture does when it runs out of evidence: prolong the last affine piece, or
emit the prior. Neither is the target. We add that the converging case
is in some respect the more dangerous, since a plausible mid-range constant causes
less suspicion than a visibly diverging ramp.

\subsection{The reach across architecture families}
\label{sec:reach}

It is tempting to treat the deficit as a symptom of insufficient capacity, and to
expect scale to resolve it. But it is instead a structural property of the
representation \citep{xu2020how}, and it propagates: an MLP sits inside nearly
every DL architecture in use, and wherever it lies on the path of a computed
quantity, that quantity inherits its limits. \S\ref{sec:composition} states the
propagation rule precisely, including the cases where it does not apply, which
are as informative as the cases where it does.

Increasing Neural Network depth does not help, and the reason is a detail of
the UAT that is often overlooked. The theorem guarantees approximation within a
compact subset $K \subset \mathbb{R}^n$ \citep{hornik1989multilayer,
cybenko1989approximation}. When affine maps are composed with standard
nonlinearities, the partition they induce is local, so additional layers increase
the density of hyperplanes inside $K$ \citep{montufar2014number}. Outside $K$ no data constrains the
optimization, and however deep or wide the Neural Network, the outermost regions remain
unbounded.

\begin{itemize}
\item \textbf{Convolutional Neural Networks (CNNs).} A convolutional layer is a sparse affine
transformation with weight sharing, followed by an activation, so a CNN is a
structurally constrained curve-fitting function across spatial dimensions. They
fail to extrapolate even to modest spatial transformations
\citep{azulay2019transformations}.
\item \textbf{Transformers.} Every Transformer block ends in a position-wise
feed-forward Neural Network, which is an MLP, and is mathematically required to
prevent the attention matrix from rank-collapsing \citep{vaswani2017attention,
dong2021attention}. Despite dynamic context routing, a Transformer remains subject
to the same region geometry and activation asymptotes at the boundary of its
learned distribution, which pushes it toward structural pattern matching rather
than compositional reasoning off-distribution \citep{dziri2023faith}.
\item \textbf{Graph Neural Networks (GNNs).} Message passing aggregates neighbourhood features
and updates node representations through MLPs \citep{kipf2017semisupervised}. The
relational inductive bias helps interpolation within known topologies, but on
graphs of unseen size or with node features outside the training manifold, the
underlying MLPs must extrapolate, with the same asymptotic consequences
\citep{yehudai2021local, xu2020how}.
\end{itemize}

\textbf{Geometric DL} deserves a separate treatment, since it is the strongest
form of the objection. By imposing invariance or equivariance to a symmetry group
$\mathfrak{G}$, it mitigates the \emph{curse of dimensionality} and improves
generalization \citep{bronstein2021geometric, cohen2016groupequivariant}. But
does operating intrinsically on manifolds overcome the asymptotic limit? It does
not, and the reason is instructive rather than dismissive: it expands the
interpolation space. A model with the added equivariance constraint no longer
operates on the original unbounded space but on the quotient
$\mathcal{X}/\mathfrak{G}$, and
inside that quotient the constraint is exact at any distance from the data,
since it was imposed as an algebraic identity and never estimated. If a shift
moves
along a direction the group does not act on, the quotient is left and the model
goes into extrapolation mode, as before. Approaches combining invariance with information
bottlenecks to bound generalization error \citep{ahuja2021invariance} inherit the
same boundary. Standard pointwise nonlinearities are defined on Euclidean
coordinates rather than intrinsically on a curved manifold, so such models apply
them in local Euclidean tangent spaces, through exponential and logarithmic maps
in hyperbolic Neural Networks \citep{ganea2018hyperbolic} and through local
coordinate charts in gauge-equivariant architectures \citep{cohen2019gauge}, and
once outside the protected quotient, the underlying MLPs converge to their
asymptotes.

This is the constructive reading we return to in \S\ref{sec:dependence}: an
imposed group provides some directions with exactness and leaves the remainder
without it, so the question becomes who determines this group.

\section{The Relational Face}
\label{sec:relational-face}

\subsection{Propositional and first-order representations}

There is a parallel limitation in relational reasoning. If asking DL
architectures to learn rules governing relational data, whether knowledge graphs,
family trees, or physical interactions, these default to propositional
representations rather
than abstract, universally quantified rules \citep{marcus2018deep,
fodor1988connectionism}.

The distinction is the classical one. A propositional system evaluates the truth
of specific ground statements. A first-order system uses variables that can be
instantiated across an unbounded domain, as in
$\forall X, Y:\ \text{Parent}(X,Y)$. Knowledge graph embeddings and GNNs
operate predominantly at the propositional level: they learn geometric vectors for
specific known entities, and therefore lack the mechanism of variable
binding, the ability to instantiate an abstract logical variable with a novel
entity at inference time \citep{greff2020binding}. Presented with an entity absent
from training, such a model has no learned vector, and deduction collapses.

\subsection{Case study: the transitive kinship discrepancy}

Consider inferring $\text{Grandparent}$ from $\text{Parent}$. The mechanism
generating the data is an exact, universally quantified rule:
\begin{equation}
\forall X, Y, Z:\ \text{Parent}(X, Y) \wedge \text{Parent}(Y, Z)
\Rightarrow \text{Grandparent}(X, Z) .
\end{equation}
Because the rule is stated over unbound variables, its truth is independent of the
entities themselves. Learn it, and it extrapolates exactly to any entities whatever.

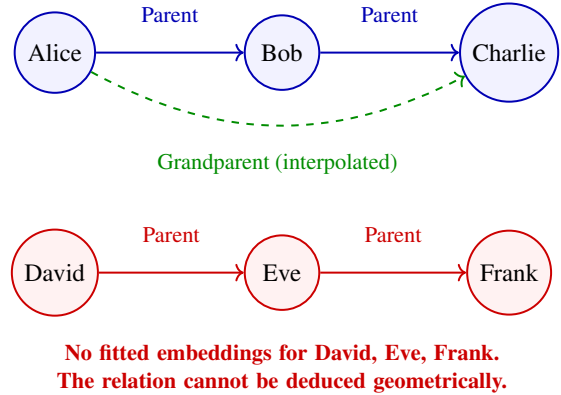
\begin{figure}[htbp]
\centering
\resizebox{0.5\textwidth}{!}{%
\begin{tikzpicture}[node distance=3.2cm,
  every node/.style={circle, draw, minimum size=1.05cm, thick, font=\small}]
\node[draw=blue!70!black, fill=blue!5] (Alice) {Alice};
\node[draw=blue!70!black, fill=blue!5] (Bob) [right of=Alice] {Bob};
\node[draw=blue!70!black, fill=blue!5] (Charlie) [right of=Bob] {Charlie};
\draw[->, thick, blue!70!black] (Alice) --
  node[above, draw=none, rectangle, font=\footnotesize] {Parent} (Bob);
\draw[->, thick, blue!70!black] (Bob) --
  node[above, draw=none, rectangle, font=\footnotesize] {Parent} (Charlie);
\draw[->, dashed, thick, green!55!black] (Alice) to[bend right=28]
  node[below, draw=none, rectangle, font=\footnotesize]
  {Grandparent (interpolated)} (Charlie);
\begin{scope}[yshift=-3.1cm]
\node[draw=red!80!black, fill=red!5] (David) {David};
\node[draw=red!80!black, fill=red!5] (Eve) [right of=David] {Eve};
\node[draw=red!80!black, fill=red!5] (Frank) [right of=Eve] {Frank};
\draw[->, thick, red!80!black] (David) --
  node[above, draw=none, rectangle, font=\footnotesize] {Parent} (Eve);
\draw[->, thick, red!80!black] (Eve) --
  node[above, draw=none, rectangle, font=\footnotesize] {Parent} (Frank);
\node[draw=none, rectangle, text=red!80!black, font=\footnotesize\bfseries,
  text width=8.4cm, align=center] at (3.2, -1.35)
  {No fitted embeddings for David, Eve, Frank.\\
   The relation cannot be deduced geometrically.};
\end{scope}
\end{tikzpicture}}
\caption{An architecture that represents an entity as a free per-entity parameter
memorizes propositional geometry and has nothing to apply to a disjoint set of
novel entities, because the rule was never represented as a first-order rule.}
\label{fig:kinship}
\end{figure}

Trained on such data, an embedding model does not learn the rule. TransE
\citep{bordes2013transe} and R-GCN \citep{schlichtkrull2018modeling} instead
learn a continuous approximation, assigning a vector to each known entity and
adjusting those vectors until the geometry agrees with the propositions observed:
\begin{align}
e_{\text{Alice}} + v_{\text{Parent}} &\approx e_{\text{Bob}} \\
e_{\text{Bob}} + v_{\text{Parent}} &\approx e_{\text{Charlie}}
\end{align}
from which $e_{\text{Alice}} + v_{\text{Grandparent}} \approx e_{\text{Charlie}}$
follows by interpolation. Introduce a disjoint family of novel entities, as in
Figure~\ref{fig:kinship}, and the arithmetic has nothing to operate on: the
vectors for the new entities were never fitted, so they carry no information about
the relation. The model did not learn the rule, it learned the geometry of the
training examples \citep{marcus2001algebraic}.

\FloatBarrier

\subsection{Propositional memorization and LLMs}

When an architecture of this kind fails to learn a rule with quantifiers it
succeeds at something else: it memorizes the propositional geometry of the
training set. When observing (Paris $\rightarrow$ France $\rightarrow$ Europe) it
assigns and fits vectors until the distances agree, and predicts that Paris is in
Europe not because it represents transitivity but because it is stored in that
geometry. Asked
about (Kyoto $\rightarrow$ Japan $\rightarrow$ Asia) with no vector for Kyoto,
the arithmetic breaks down.

Because LLMs generate fluent text they project
the impression of an internal symbolic engine, when they are autoregressive
predictors over token co-occurrence \citep{bender2020climbing}. Answering a
syllogism correctly is not executing $\text{Human}(X) \Rightarrow
\text{Mortal}(X)$; it is predicting a highly probable token sequence whose
structure is dense in the training distribution. When the semantics of a puzzle
are inverted, or entities are replaced with novel synthetic nouns, performance
degrades sharply \citep{dziri2023faith}, and
\citet{mirzadeh2024gsmsymbolic} reach the same conclusion from perturbed
mathematical word problems. \citet{lee2025loth} reach it from the other side, by
scoring the process rather than the answer on ARC-AGI: the models that reproduce a
correct output grid frequently fail the intermediate steps that would have produced it,
so the output is not evidence that a rule was applied. Such models learn what logic
looks like rather than the true rules of logic.

\subsection{The limits of Turing-complete neural architectures}

Neural architectures explicitly designed to simulate algorithmic reasoning are
provably Turing complete, and theoretical capacity is not empirical realization. Because
these designs remain fully continuous and differentiable, they do not escape the
representational limit.

\begin{itemize}
\item \textbf{Memory-augmented networks.} Neural Turing Machines
\citep{graves2014neural} and Differentiable Neural Computers
\citep{graves2016hybrid} decouple computation from memory, which is the right
move. To stay end-to-end differentiable they cannot use discrete pointers, and
rely instead on continuous addressing driven by similarity. Introduce an entity
outside the training distribution and the similarity search has no reliable
target.
\item \textbf{Adaptive computation and Universal Transformers.} Halting
computation dynamically \citep{graves2016adaptive, dehghani2019universal} achieves
length generalization, and the loop applies the same continuous affine
transformations recursively, so abstract relational generalization does not
follow from it.
\item \textbf{Relational Networks.} Imposing a structural prior by evaluating
every pair $(x_i, x_j)$ \citep{santoro2017simple} does not change what processes
each pair, which is a standard MLP receiving coordinates outside its empirical
region.
\end{itemize}

The pattern is the one already stated, and it is the compositional rule of
\S\ref{sec:composition} seen from the relational side: Turing completeness
describes what an architecture could compute given the right weights, and says
nothing about what lies on the inference path of any particular quantity. In each
of these designs the quantity that must generalize to a novel entity has a fitted
similarity computation, a fitted addressing mechanism, or a fitted inner Neural Network on
its route, so it inherits the approximation limits accordingly.

\subsection{Stated at the right scope}
\label{sec:right-scope}

The previously described failure belongs to architectures that represent an
entity as a free parameter fitted to that entity. It must not be claimed that
neural architectures cannot bind variables. Two published systems show they can,
and both are instructive.

\textbf{ESBN} \citep{webb2021esbn} augments a recurrent controller with an external
memory whose key column holds abstract variables and whose value column holds
perceptual embeddings. The controller never receives entity embeddings directly;
it emits keys and receives back retrieved keys and confidences, so the
rule-learning pathway is structurally entity-independent. It attains near-perfect
generalization of learned rules to novel entities. Its scope is narrow: its tasks, same/different, relational match-to-sample,
distribution-of-three and identity rules, concern relations among items presented
together within a single problem, not multi-step chained inference binding an
intermediate variable across steps over an open domain of constants. The
grandparent rule example we gave is of the latter kind.

\textbf{GraIL} \citep{teru2020grail} reasons over subgraph structure with node
labelling relative to the target nodes, making the representation
entity-independent, and generalizes to unseen entities and entire unseen graphs.
Its own theoretical claim is to represent ``a useful subset of first-order
logic.''

Neither refutes the representational thesis. Both confirm it, and this is the
cleanest evidence available for it. ESBN binds variables because indirection was
built into its representation. If one removes the indirection the capability goes
with it. GraIL generalizes to unseen entities because entity identity was designed out
of what it represents. In each case the generalization capability arrived exactly
when the alternative representation did, which is our thesis, and \S\ref{sec:composition} says why in
architectural terms: both designs work by removing a fitted component from the
inference path of the quantity that had to generalize. And \citet{barcelo2020logical} give the
precise version for GNNs, a fragment of FOL with counting
rather than propositional logic, so the real limit consists of a
representational boundary and not an unsurpassable wall. The binding problem
persists in frontier multimodal systems and is documented by
\citet{campbell2024binding}, from authors who elsewhere advocate architectural
solutions to it.

GraIL also draws the distinction this paper is about. It is entity-independent, it
generalizes, but it still scores candidate triples with a function approximation
rather than executing a relation. Its generalization is empirical and graded. Better
representability bought better generalization without buying exactness, so exact
generalization is still out of reach for this system.

\section{Exactness at Inference}
\label{sec:criterion}

\begin{criterion}[Exactness at inference]
\label{crit:main}
A system generalizes outside its training distribution on a given input only if
the representation it consults on that input is \emph{structurally equivalent}
to the underlying data-generating mechanism, so that the operation performed
\emph{computes} the intended function or relation, rather than evaluating a
continuous surrogate with the wrong representation, whose agreement with that
function or relation was calibrated on a region of input space that the given
input does not lie in.
\end{criterion}

\paragraph{What structural equivalence requires.} We mean the previous
definition in the algebraic sense rather than the physical one, since the
objection that a model executes arithmetic on silicon and not on the mechanism
itself is correct but beside the point. A representation is structurally
equivalent to a mechanism when its constituents and its composition operations
map onto the mechanism's in a way that preserves the mechanism's relations on
every admissible input, so that agreement holds by construction over the
whole domain and not by calibration on a sampled part of it. This is the sense
of universal algebra and model theory, a homomorphism of structures that
preserves the operations and relations. Floating-point
arithmetic evaluating $x^{2}$ is structurally equivalent to squaring, to within
a rounding error that is a property of the number system and not of the data;
resolution over a clause is structurally equivalent to the quantified rule the
clause states; a partition of affine pieces fitted to samples of a curved law is
not structurally equivalent to that law, however small the error on the samples
can be.
The definition also makes the criterion's scope plain: equivalence is to the
mechanism as the problem presents it, so an encoding that is itself wrong cannot
be rescued by exact execution downstream of it, a point we return to in
\S\ref{sec:dependence}.

Three clarifications are still needed.

\subsection{Exactness is not discreteness}
\label{sec:not-discreteness}

Discreteness is what exactness reduces to when the intended semantics is classical
logic, whose ground atoms and truth values are discrete. Discreteness is not the
right criterion, and conflating them misclassifies systems in both directions.

In one direction, an exact operation may be continuous-valued. A sum-product
tensor contraction computing an exact marginal has codomain $[0,1]$, has nothing
fitted in it, and extrapolates for the same reason $f(x)=x^2$ does: it computes
the intended function rather than a surrogate calibrated on data.
Figure~\ref{fig:contraction} shows the operation in both of its readings. A system may therefore reason under uncertainty and satisfy Criterion \ref{crit:main},
which matters for any domain where the evidence is partial.

\begin{figure}[htbp]
\centering
\begin{tikzpicture}[
  font=\footnotesize,
  cell/.style={draw=black!60, minimum size=0.40cm, inner sep=0pt, font=\scriptsize, anchor=center},
  one/.style={cell, fill=black!12},
  box/.style={draw, rounded corners=1.5pt, minimum width=0.62cm, minimum height=0.46cm, fill=black!5, font=\scriptsize},
  leg/.style={thick},
  lbl/.style={font=\scriptsize, inner sep=1pt}
]
\begin{scope}
\node[font=\footnotesize\bfseries, anchor=west] at (-0.55,2.45) {(a) Boolean tensors, $T=0$: the Grandparent rule};
\matrix (P1) [matrix of nodes, nodes={cell}, column sep=-\pgflinewidth, row sep=-\pgflinewidth] at (0.45,1.15) {
  0 & |[one]| 1 & 0 \\ 0 & 0 & |[one]| 1 \\ 0 & 0 & 0 \\ };
\node[lbl, above=1pt of P1] {$P_{xy}$};
\node[lbl, left=1pt of P1-1-1] {A}; \node[lbl, left=1pt of P1-2-1] {B}; \node[lbl, left=1pt of P1-3-1] {C};
\node[lbl] at (1.55,1.15) {$\cdot$};
\matrix (P2) [matrix of nodes, nodes={cell}, column sep=-\pgflinewidth, row sep=-\pgflinewidth] at (2.6,1.15) {
  0 & |[one]| 1 & 0 \\ 0 & 0 & |[one]| 1 \\ 0 & 0 & 0 \\ };
\node[lbl, above=1pt of P2] {$P_{yz}$};
\node[lbl] at (3.75,1.15) {$\xrightarrow{\;\sum_{y}\;}$};
\matrix (G) [matrix of nodes, nodes={cell}, column sep=-\pgflinewidth, row sep=-\pgflinewidth] at (4.95,1.15) {
  0 & 0 & |[one]| 1 \\ 0 & 0 & 0 \\ 0 & 0 & 0 \\ };
\node[lbl, above=1pt of G] {$G_{xz}$};
\node[lbl, below=1pt of P1-3-1] {A}; \node[lbl, below=1pt of P1-3-2] {B}; \node[lbl, below=1pt of P1-3-3] {C};
\node[lbl, below=1pt of P2-3-1] {A}; \node[lbl, below=1pt of P2-3-2] {B}; \node[lbl, below=1pt of P2-3-3] {C};
\node[lbl, below=1pt of G-3-1] {A}; \node[lbl, below=1pt of G-3-2] {B}; \node[lbl, below=1pt of G-3-3] {C};
\node[lbl, left=1pt of P2-1-1] {A}; \node[lbl, left=1pt of P2-2-1] {B}; \node[lbl, left=1pt of P2-3-1] {C};
\node[lbl, left=1pt of G-1-1] {A}; \node[lbl, left=1pt of G-2-1] {B}; \node[lbl, left=1pt of G-3-1] {C};
\node[lbl, right=1pt of G-1-3] {A$\to$C};
\node[box] (b1) at (0.9,-0.75) {$P$};
\node[box] (b2) at (2.7,-0.75) {$P$};
\draw[leg] (b1.east) -- node[lbl, above] {$y$} (b2.west);
\draw[leg] (b1.west) -- ++(-0.5,0) node[lbl, left] {$x$};
\draw[leg] (b2.east) -- ++(0.5,0) node[lbl, right] {$z$};
\node[lbl, text width=6.6cm, align=left, anchor=north west] at (-0.55,-1.2)
  {$G_{xz}=\sigma\big(\textstyle\sum_{y}P_{xy}P_{yz}\big)$ over $\{$A, B, C$\}$: the shared index $y$ is summed out and the step function at $T=0$ keeps the values in $\{0,1\}$. The incidence tensor of Grandparent is computed, not fitted.};
\end{scope}
\begin{scope}[xshift=7.6cm]
\node[font=\footnotesize\bfseries, anchor=west] at (-0.55,2.45) {(b) Real tensors: an exact marginal};
\node[box] (px) at (0.3,1.15) {$p(x)$};
\node[box, minimum width=0.95cm] (pyx) at (1.75,1.15) {$p(y\,|\,x)$};
\node[box, minimum width=0.95cm] (pzy) at (3.35,1.15) {$p(z\,|\,y)$};
\draw[leg] (px.east) -- node[lbl, above] {$x$} (pyx.west);
\draw[leg] (pyx.east) -- node[lbl, above] {$y$} (pzy.west);
\draw[leg] (pzy.east) -- ++(0.5,0) node[lbl, right] {$z$};
\node[lbl] at (2.0,0.45) {$p(z)=\sum_{x}\sum_{y}\,p(x)\,p(y\,|\,x)\,p(z\,|\,y)$};
\node[lbl, text width=6.4cm, align=left, anchor=north west] at (-0.55,0.05)
  {Same operation, same diagram, entries in $[0,1]$. The contraction of the chain's factors is the exact marginal of $z$: its codomain is continuous, nothing in it is fitted, and it holds for every input for the same reason $f(x)=x^{2}$ does. A box is a tensor, a line an index, a joined line a summed index, an open line an index of the result. Reading (a) is Datalog; reading (b) is a PGM. The criterion is satisfied in both, since what is executed is the relation itself.};
\end{scope}
\end{tikzpicture}
\caption{A sum-product tensor contraction in its two readings. (a) With Boolean
incidence tensors and a step function, the contraction over the shared index
executes the Grandparent rule of \S\ref{sec:relational-face} exactly, over the
closed domain the indices enumerate. (b) With the same shape over real factors,
it computes an exact marginal with codomain $[0,1]$. Discreteness is a property
of the entries; exactness is a property of the operation, and the operation is
the same in both panels.}
\label{fig:contraction}
\end{figure}

In the other direction, a discrete-valued operation may be inexact. A standard
Neural Network whose output is thresholded to a hard label emits a discrete value,
but this threshold is a fitted boundary; hardening it changes the codomain
and not the epistemic status. Exactness is a property of what is computed.
Discreteness is a property of the codomain. The distinction that survives is
between the function computed and the substrate computing it, and it is the
function that must be exact.

One qualification applies throughout. Exactness is always exactness relative
to the object being executed: an exact marginal is exact given the factors, and a
logic program is exact given the clauses. Where those were themselves induced, the
induction remains fallible. The claim is not that induction is infallible but that
its fallibility and the execution's exactness are separable, which is what makes
the error localizable.

\subsection{It is a property of inference, not of training}

Nothing in Criterion \ref{crit:main} constrains how the right structure can be
found. A system may search by gradient descent, by relaxation, by enumeration or
by sampling, and satisfy the criterion provided that what it consults at
inference time is the resulting structure and not the search mechanism. This is
why the criterion does not imply avoiding Neural Networks, and why it allows for
the whole of neurally guided search approaches.

The consequence is a qualitative difference in how systems fail. Where an
approximation is consulted at inference, failure is graded: it degrades as
inputs move away from the data, continuously and without a signal. Where a
structure is executed, failure is discrete: if the wrong structure was induced,
then the structure is wrong everywhere in the same way, which is inspectable and
which the induction can be blamed for. Fallible induction with exact execution is
a different epistemic situation from fallible execution, and only the first
localizes the error at training time, as will be shown in \S\ref{sec:no-signal}.

\subsection{Composition: containing an MLP, and inheriting its limits}
\label{sec:composition}

Almost every current DL architecture uses, somewhere inside itself, a standard
MLP. The Transformer's position-wise feed-forward block is one,
and is mathematically required to keep attention from rank-collapsing
\citep{dong2021attention}. A message-passing Neural Network updates node states
through one \citep{kipf2017semisupervised}. A Relational Network processes every
object pair with one \citep{santoro2017simple}. An LTN grounds each predicate in
one \citep{badreddine2022logictensor}. A scalar-invariant model applies one to
the invariants it computes \citep{villar2021scalars}. The temptation is to
conclude that all of them therefore inherit the limits of
\S\ref{sec:continuous-face}, but this would be an unqualified version of our
criterion.

Three cases in this paper refute the unqualified version. As mentioned before, an
ESBN controller is built from standard components and nonetheless binds variables
to novel entities \citep{webb2021esbn}. A scalar-invariant model is an MLP applied
to $r$, and it is exact along every direction its group acts on. A Hamiltonian
Neural Network's Hamiltonian is an arbitrary fitted Neural Network, and energy is
still conserved exactly \citep{greydanus2019hamiltonian}. There is a further
technical objection: an MLP receiving normalized activations is never evaluated
far outside its own training domain, because it is constrained by the
normalization independently of the Neural Network's input, which is the same
mechanism that produces convergence rather than divergence in \S\ref{sec:scope}.
Presence of a fitted component is evidently not sufficient for the inheritance of
MLP limits.

The correct statement is the compositional form of Criterion \ref{crit:main}, and
it is a statement about paths rather than about components.

\begin{quote}
\emph{A computed quantity inherits the limits of every fitted component lying on
its inference path. If the fitted component is not in its inference path, it
constrains other quantities and not this one.}
\end{quote}

This resolves all the cases at once. The Transformer's feed-forward block lies
on the path of the output, so the output inherits its limits. The message-passing
update lies on the path of every node representation, so the nodes inherit the
limits. The invariant model's MLP lies on the path from $r$ to its prediction but
on no path from angular position to anything, since the angle is discarded before
the MLP is reached, which is why such a model is simultaneously exact on the
group's orbits and fitted along the law, a dissociation observable directly in a
controlled setting (\S\ref{sec:boundary-bias}). ESBN's indirection is precisely a device for keeping
entity embeddings off the controller's path, and the binding capability it buys is
the capability of the quantity that no longer has a fitted component on its route.
In a Hamiltonian Neural Network the symplectic integrator lies on the path of energy
conservation while the fitted Neural Network lies on the path of the dynamics, and the
two properties behave accordingly: conservation is exact, and the trajectory is
not.

The criterion has a quantitative aspect that matters where modules are chained.
Exactness is preserved under composition and approximation is not. A module that
computes its relation exactly contributes a residual of zero, and zero is a
fixed point of composition, whereas a fitted module contributes a residual
$\varepsilon > 0$ that every later stage transforms rather than removes. Across
$n$ stages with downstream Lipschitz constants $L_{k}$ the accumulated error is
bounded by $\sum_{k \le n} \varepsilon_{k} \prod_{j > k} L_{j}$, a bound
derived in \S\ref{sec:derivations}, which grows
geometrically in $n$ wherever the composed map expands, so a residual of
$10^{-4}$ that is invisible at one hop is not invisible at twenty. This is the
mechanism behind a familiar observation, that architectures with excellent
single-step accuracy degrade on multi-hop reasoning and on long rollouts while
their one-step error in distribution looks perfect, and it is the reason the
composition evidence of \S\ref{sec:arc} favours systems whose stages are exact:
what they compose is a residual that stays at zero.

Two consequences follow. The first is diagnostic. To predict where an
architecture will fail OOD, one does not ask whether it contains a
fitted component but which quantities have one on their path. That question is answerable by inspection of the architecture, before any
experiment, and it is what makes the criterion useful rather than merely correct.
The second is that placing a Neural Network off the inference path is a design move
available to anyone, and it is the same move \S\ref{sec:heterogeneous} recommends
for search guidance: an MLP-based model that ranks candidate structures influences which
hypothesis is selected without being on the inference path.

The unqualified reading is nonetheless right for the general case, and the
exceptions usually share the same feature. This feature is that something was
deliberately built to keep the fitted component off a path, and someone had to
know which path to protect. That is \S\ref{sec:dependence}'s point arriving from
another direction: architectures are exact where a designer arranged for them to
be, and fitted everywhere else, which corresponds to the totality of standard
Neural Network architectures.

\subsection{Sorting neuro-symbolic systems}
\label{sec:sorting}

The failure of the architectures above reveals an apparent paradox: the very
mechanism that allows Neural Networks to learn end-to-end, differentiability, is what
prevents them from executing abstract logic. To compute a gradient the loss
landscape must be continuous and smooth, but to execute logic reliably the
operation performed must be exact rather than approximate. This becomes clear on
differentiable neuro-symbolic systems, and our Criterion~\ref{crit:main} sorts
them in a way that a surface taxonomy of neuro-symbolic systems
\citep{kautz2022third}, which classifies them by how the neural and the symbolic
components are wired together, does not.

\paragraph{Logic Tensor Network fails it.} LTNs unify connectionism and logic by
relaxing FOL into a continuous domain using fuzzy t-norms
\citep{badreddine2022logictensor, garcez2023neurosymbolic}. Logical rules can then
act as differentiable loss functions during training, and the predicates remain
parameterized by Neural Networks. Crucially no standalone discrete rule is
extracted. To evaluate a relation between novel entities at inference, an LTN must
pass them through its continuous layers, which on unseen inputs produces an
asymptotic geometric artefact, and the fuzzy operators aggregate those arbitrary
continuous outputs into a confident truth value. Thresholding would not repair it
because the predicates are fitted Neural Networks, so a hard decision boundary is
still a fitted, approximate boundary. Note also that on interior values the
product t-norm equals the probability of a conjunction only under independence,
which generally fails, so a composed formula is a surrogate for both the Boolean
reading and the probabilistic reading, and an instance of neither. Two qualifications are owed: LTN defines
a proof-by-refutation mode we do not analyse here, and practitioners often apply a
post-hoc threshold at $0.5$, which changes the output's type without changing what
produced it.

\paragraph{Differentiable ILP passes it.} $\partial$ILP \citep{evans2018learning}
appears to contradict the requirement but it does not. The
continuous relaxation is used only during training, to smooth a
combinatorial search space; weights are regularized toward $\{0,1\}$; at inference
the system extracts a strictly discrete symbolic rule and discards the Neural Network
entirely. It succeeds precisely because it abandons the continuous manifold before
inference, using differentiability as a temporary search heuristic to discover a
discrete extrapolative structure. Evans and Grefenstette report the signature of
exactness directly: trained on integers from $0$ to $6$, ``the learned program
works effectively on test integers of any size,'' with ``no chance of the learned
program suddenly failing when the integers reach a certain size.''

\paragraph{Tensor Logic at zero temperature passes it.} This case is instructive
because it first appears to be a counterexample. Tensor Logic
\citep{domingos2025tensorlogic} proposes a language whose unifying construct is
the tensor equation, resting on the observation that a logical rule and an
Einstein summation are essentially the same operation, so that neural and
symbolic computation are dual aspects of the same construct. Reasoning runs over learned
embeddings throughout, with no separate artefact written out and no Neural Network
discarded, which on the surface groups it with LTNs. Its behaviour is the
opposite, and a temperature parameter makes the difference. Applying
$\sigma(x,T) = 1/(1+e^{-x/T})$ to each equation, Domingos observes that $T=0$
``effectively reduces the Gram matrix to the identity matrix, making the program's
reasoning purely deductive,'' and that the system is ``immune to hallucinations at
sufficiently low temperature.'' Temperature is settable per rule, so rules
expressing mathematical truths may run at $T=0$ while rules accumulating weak
evidence may run high.

Our reading is that at $T \to 0$ the sigmoid applied to every equation
approaches a step function. Consequently, the inferred tensors become Boolean and
the computed function becomes discrete, which is necessary for classical logic as
noted in \S\ref{sec:relational-face}, even though the underlying arithmetic
remains continuous.

This zero-temperature limit is not merely a discretization. A Gram matrix equal
to the identity implies that the embeddings are orthonormal, functioning as a
one-hot encoding up to an unobservable rotation: any orthonormal family is the
image of the standard basis under an orthogonal transformation, and a
contraction sees its embeddings only through inner products, which such a
transformation leaves unchanged. The computational carrier, dense real tensors
contracted by real multiplications and sums, remains entirely continuous, with
no separate logical rule store or symbol table. Only the values become discrete.
A Boolean tensor indexed by an orthonormal family is simply the incidence tensor
of a finite relation, that is, the table of which tuples of index values stand
in the relation. This explains why the mathematical equivalence reaches Datalog
and no further, and why introducing a novel entity requires a new
index dimension rather than just a new constant.

The symbolic content is natively present because the tensor equation, written in
contraction syntax, already serves as the symbolic program. Unlike $\partial$ILP,
which searches via continuous relaxation but must extract a discrete rule and
discard the Neural Network, Tensor Logic collapses the relaxation in place. The
difference between the two is implementation, not semantics.

This physical implementation imposes strict geometric boundaries on the tensors.
Because the system computes via fixed-dimensional linear algebra, it cannot
natively represent infinite functional recursion: a tensor of fixed shape
enumerates a fixed set of index values, whereas a function symbol builds new
terms without bound, and those terms have no index to range over. It therefore
achieves equivalence only with Datalog, which uses finite constants and
function-free terms, and not with full Prolog, which requires an infinite
Herbrand universe and terms of unbounded depth. Also, this equivalence holds only under two familiar conditions for finite and complete Datalog
evaluation \citep{abiteboul1995foundations}: the rules must be range-restricted,
and each index must cover the active domain. Attempting to dynamically expand
dense tensors to bind novel entities multiplies costs across the
remaining dimensions and shatters the static execution graphs that accelerators
compile once and reuse \citep{jouppi2017tpu}.

Consequently, while the in-place tensor is exact over a known, closed domain,
binding variables to novel entities requires abandoning the strict in-place
constraint. The system must either use external-memory indirection to allocate
new symbols without reshaping the core reasoning tensors \citep{graves2016hybrid,
webb2021esbn}, or explicitly extract the rule for a standard logic engine to
execute, as $\partial$ILP does \citep{evans2018learning}. Both approaches align
with the propagation rule of \S\ref{sec:composition} by keeping a fitted
representation of the entity off the inference path. Conversely, if $T > 0$, the
continuous embedding geometry carries inferential weight, entity representations
revert to fitted surrogates, and the relational limits of
\S\ref{sec:relational-face} fully return inside the closed domain. Per-rule
temperature therefore acts as a declaration of which parts of a model hold
exactly and which remain approximate. Temperature above zero is a relaxation of
the logical reading, in which the embedding geometry rather than the relation
carries the inference, and it is not the same thing as probabilistic inference.
Tensor Logic also executes the latter, as a sum-product contraction over factor
tensors rather than a softened contraction over embeddings, and that regime can
be exact in its own semantics; \S\ref{sec:graded} says under which conditions.

One qualification applies to both passing cases: the exactness of the execution
depends on a fallible induction. Tensor Logic's exactness relies on the learned
embeddings being near-orthogonal, just as $\partial$ILP's exactness relies on
accepting the specific extracted clause. In both systems the induction is
fallible, but the execution remains exact. These cases therefore corroborate our criterion: even when beginning from the premise that logic is fundamentally a
continuous Einstein summation, reliable logical reasoning still requires the
zero-temperature limit, with exact probabilistic reasoning available under the
different semantics of \S\ref{sec:graded}.

A fully differentiable system achieving high zero-shot accuracy on novel
entities does not refute this sorting. Our criterion concerns what is computed,
not what is scored. High accuracy on a specific test distribution shift does not
guarantee structural exactness. To falsify this criterion, an unthresholded
continuous system would need to demonstrate exact, non-degrading agreement on an
adversarially chosen distribution shift. We are not aware of any such demonstration.

\subsection{Graded semantics without surrendering exactness}
\label{sec:graded}

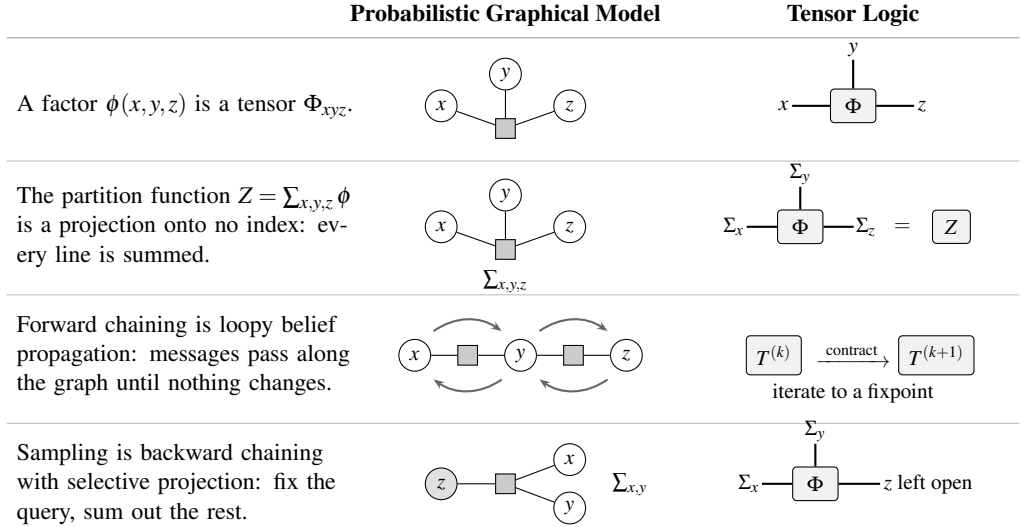
\begin{figure}[htbp]
\centering
\begin{tikzpicture}[
  font=\footnotesize,
  var/.style={draw, circle, minimum size=0.42cm, inner sep=0pt, font=\scriptsize},
  qvar/.style={var, fill=black!12},
  fac/.style={draw, fill=black!20, minimum size=0.26cm, inner sep=0pt},
  box/.style={draw, rounded corners=1.5pt, minimum width=0.6cm, minimum height=0.44cm, fill=black!5, font=\scriptsize},
  leg/.style={thick},
  sum/.style={font=\scriptsize, inner sep=0.5pt},
  lbl/.style={font=\scriptsize, inner sep=1pt},
  rowlbl/.style={font=\footnotesize, anchor=west, text width=4.5cm, align=left},
  msg/.style={-{Stealth[length=4pt]}, thick, black!60}
]
\def\rowA{2.35}\def\rowB{0.75}\def\rowC{-0.95}\def\rowD{-2.65}
\node[font=\footnotesize\bfseries] at (6.6,3.55) {Probabilistic Graphical Model};
\node[font=\footnotesize\bfseries] at (11.2,3.55) {Tensor Logic};
\draw[black!40] (0,3.25) -- (13.4,3.25);
\node[rowlbl] at (0,\rowA) {A factor $\phi(x,y,z)$ is a tensor $\Phi_{xyz}$.};
\node[var] (r1x) at (5.75,\rowA) {$x$}; \node[var] (r1y) at (6.6,\rowA+0.42) {$y$}; \node[var] (r1z) at (7.45,\rowA) {$z$};
\node[fac] (r1f) at (6.6,\rowA-0.3) {}; \draw (r1f) -- (r1x); \draw (r1f) -- (r1y); \draw (r1f) -- (r1z);
\node[box] (r1t) at (11.2,\rowA) {$\Phi$};
\draw[leg] (r1t.west) -- ++(-0.5,0) node[lbl,left] {$x$};
\draw[leg] (r1t.north) -- ++(0,0.38) node[lbl,above] {$y$};
\draw[leg] (r1t.east) -- ++(0.5,0) node[lbl,right] {$z$};
\draw[black!25] (0,1.62) -- (13.4,1.62);
\node[rowlbl] at (0,\rowB) {The partition function $Z=\sum_{x,y,z}\phi$ is a projection onto no index: every line is summed.};
\node[var] (r2x) at (5.75,\rowB) {$x$}; \node[var] (r2y) at (6.6,\rowB+0.42) {$y$}; \node[var] (r2z) at (7.45,\rowB) {$z$};
\node[fac] (r2f) at (6.6,\rowB-0.3) {}; \draw (r2f) -- (r2x); \draw (r2f) -- (r2y); \draw (r2f) -- (r2z);
\node[lbl] at (6.6,\rowB-0.72) {$\sum_{x,y,z}$};
\node[box] (r2t) at (10.5,\rowB) {$\Phi$};
\draw[leg] (r2t.west) -- ++(-0.4,0) node[sum,left] {$\Sigma_x$};
\draw[leg] (r2t.north) -- ++(0,0.3) node[sum,above] {$\Sigma_y$};
\draw[leg] (r2t.east) -- ++(0.4,0) node[sum,right] {$\Sigma_z$};
\node[lbl] at (11.85,\rowB) {$=$};
\node[box, minimum width=0.5cm] at (12.5,\rowB) {$Z$};
\draw[black!25] (0,-0.15) -- (13.4,-0.15);
\node[rowlbl] at (0,\rowC) {Forward chaining is loopy belief propagation: messages pass along the graph until nothing changes.};
\node[var] (r3a) at (5.4,\rowC) {$x$}; \node[fac] (r3f1) at (6.1,\rowC) {}; \node[var] (r3b) at (6.8,\rowC) {$y$}; \node[fac] (r3f2) at (7.5,\rowC) {}; \node[var] (r3c) at (8.2,\rowC) {$z$};
\draw (r3a)--(r3f1)--(r3b)--(r3f2)--(r3c);
\draw[msg] (5.65,\rowC+0.32) to[bend left=35] (6.55,\rowC+0.32);
\draw[msg] (7.05,\rowC+0.32) to[bend left=35] (7.95,\rowC+0.32);
\draw[msg] (7.95,\rowC-0.32) to[bend left=35] (7.05,\rowC-0.32);
\draw[msg] (6.55,\rowC-0.32) to[bend left=35] (5.65,\rowC-0.32);
\node[box] (r3t1) at (10.15,\rowC) {$T^{(k)}$};
\node[lbl] at (11.2,\rowC) {$\xrightarrow{\ \text{contract}\ }$};
\node[box] (r3t2) at (12.3,\rowC) {$T^{(k+1)}$};
\node[lbl] at (11.2,\rowC-0.5) {iterate to a fixpoint};
\draw[black!25] (0,-1.85) -- (13.4,-1.85);
\node[rowlbl] at (0,\rowD) {Sampling is backward chaining with selective projection: fix the query, sum out the rest.};
\node[qvar] (r4q) at (5.75,\rowD) {$z$}; \node[fac] (r4f) at (6.6,\rowD) {}; \node[var] (r4x) at (7.45,\rowD+0.32) {$x$}; \node[var] (r4y) at (7.45,\rowD-0.32) {$y$};
\draw (r4q)--(r4f); \draw (r4f)--(r4x); \draw (r4f)--(r4y);
\node[lbl] at (8.25,\rowD) {$\sum_{x,y}$};
\node[box] (r4t) at (10.7,\rowD) {$\Phi$};
\draw[leg] (r4t.west) -- ++(-0.4,0) node[sum,left] {$\Sigma_x$};
\draw[leg] (r4t.north) -- ++(0,0.3) node[sum,above] {$\Sigma_y$};
\draw[leg] (r4t.east) -- ++(0.55,0) node[lbl,right] {$z$ left open};
\end{tikzpicture}
\caption{The correspondence Tensor Logic provides between PGM operations and
tensor operations \citep{domingos2025tensorlogic}. Each row pairs the graphical
object on the left with the contraction that computes it on the right, in the
diagram notation of Figure~\ref{fig:contraction}: a box is a tensor, a line an
index, and a line closed by $\Sigma_i$ an index $i$ summed out. Under this
correspondence a marginal is a contraction with one line left open, which is why
an exact marginal is continuous-valued and exact at the same time.}
\label{fig:pgm-tensor}
\end{figure}

One clarification prevents reading the criterion too narrowly.
Discreteness is what exactness reduces to when the intended semantics is classical
logic, and classical logic is not the only semantics available. Tensor Logic can also implement Probabilistic Graphical Models (PGMs) \citep{koller2009probabilistic}
directly, with a factor as a tensor, the partition function as a projection,
forward chaining as loopy belief propagation, and sampling as backward chaining
with selective projection (Figure~\ref{fig:pgm-tensor}). A sum-product tensor
contraction computing an exact marginal is exact and continuous-valued at the same
time: its codomain is $[0,1]$, nothing in it is fitted, and it extrapolates correctly for
the same reason $f(x)=x^2$ does. A system may therefore reason under uncertainty
without surrendering anything our criterion requires, which matters wherever the evidence is partial or noisy. Probabilistic logic programming makes the
same point from the logic side. ProbLog \citep{deraedt2007problog} computes the
exact probability of a query under the distribution semantics, so its inference is
an exact operation with codomain $[0,1]$, and DeepProbLog
\citep{manhaeve2018deepproblog} keeps that inference exact while letting neural
predicates supply the probabilities it operates on, which is the division of
labour the propagation rule of \S\ref{sec:composition} describes: the logical
inference is exact relative to the probabilities the neural predicates emit, but
inherits their limits on the quantities that pass through them, in this case the probabilities themselves. A neural predicate is a fitted
estimator of a conditional probability, and being a statistical estimate does not
exempt it from the criterion: the probability it emits is a function of its input
like any other, approximated rather than computed, so outside its training support
it is subject to the same limits as any other fitted output.

This is also the sharpest way to separate the three presented neuro-symbolic
architectures on uncertainty, and the ordering may not be intuitive. An LTN
appears built for uncertain environments and is not exact under any semantics: a
t-norm agrees with Boolean
conjunction on $\{0,1\}$, but an LTN never operates there, and on interior values
the product t-norm equals the probability of a conjunction only when the conjuncts are probabilistically independent, $P(a\wedge b)=P(a)\,P(b)$, an assumption that many real-world applications violate.
$\partial$ILP's relaxation is a device for searching a discrete space rather than a choice of
semantics for uncertainty, since the graded values it carries express confidence
about which clause is correct, not uncertainty about the world those clauses
describe, and what it delivers is a Boolean program. It therefore performs no
probabilistic inference in ProbLog's sense: no distribution over possible worlds is
defined, and no query is assigned a probability. Tensor Logic is the only one
of the three in which graded reasoning is a semantic choice rather than a relaxation,
which is what per-rule temperature makes explicit, holding mathematical truths at
$T=0$ and evidence accumulation with $T>0$, within a single program. Whether Tensor
Logic reasons exactly under uncertainty, in the way ProbLog does, therefore has a
definite answer: in its PGM reading it computes an exact marginal relative to
the factors it is given, under two conditions. The contraction must be carried
out exactly, since, on a graph with cycles, loopy belief propagation approximates
the marginal, and an approximated marginal is a surrogate whatever its codomain.
But to fulfil our criterion, the factors must be given rather than fitted on the
inference path, which is the same condition DeepProbLog's neural predicates respect. The
temperature relaxation of \S\ref{sec:sorting} is a third thing, a softened
logical reading, and it is not exact under either semantics. The two readings
should not be confused: the PGM reading involves no temperature at all, and its
exactness is governed by the two conditions just stated. A rule executed at $T>0$
therefore never meets the criterion, in an uncertain, noisy or partially observed
environment as much as in any other, unless the softened similarity it computes is
itself the generating mechanism. Where the evidence is uncertain, exactness is available through the PGM reading over given factors, while raising the temperature does not provide it.

The approach we proposed in previous work
\citep{rocha2020overcomingrl, rocha2020towardsagi, rocha2024programsynthesis}, and
to which we return in \S\ref{sec:follows}, inducing first-order clauses executed by a
resolution engine, secures exactness by the most direct route available, which is
to induce the discrete structure and run it. This is a proposed design choice,
also made for tractability and inspectability, but is not the only route
available.

\section{The Epistemic Corollary: Competence Is a Property of the Hypothesis Class}
\label{sec:epistemic}

The application of our criterion so far was concerned with where a model is
correct outside its training data. But it has a second consequence, which concerns
where a model can know whether or not it is correct outside of its training data, and the
two boundaries turn out to coincide. The prevailing treatment
of not-knowing is additive: train a Neural Network, then append an uncertainty
estimator computed from the fitted model or from the geometry of its training data
rather than from a hypothesis (an ensemble's disagreement
\citep{lakshminarayanan2017ensembles}, a distance score in input or feature space
\citep{lee2018simple, sun2022knn}), and abstain where the estimator sounds the
alarm. Our criterion explains why this
cannot work in general. Abstention is a statement about what the training data
determines and does not determine, and determination is always relative to a
hypothesis class and its constraints: the same training data that leaves a query
open under one constraint may fix it completely under another. A model that consults no exact hypothesis at inference has nothing
definite to abstain from, and no estimator bolted on afterwards can supply
the structural bounds the hypothesis class itself lacks. This section makes that
argument concrete on a single experiment.

\subsection{A fitted model carries no internal signal of recovery}
\label{sec:no-signal}

\begin{figure}[htbp]
\centering
\includegraphics[width=0.24\linewidth]{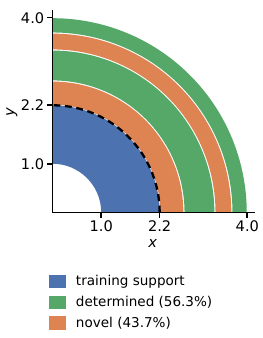}
\caption{Training annulus ($r\in[1,2.2]$) and radial query region
($r\in[2.2,4]$) on the same wedge. Periodicity in $u=x^{2}+y^{2}$ folds
alternating query bands back inside the training $u$-support (determined,
$56.3\%$ in aggregate) or leaves them outside it (novel, $43.7\%$). Membership
is fixed by radius but alternates in bands, so it is not monotone in distance
from the training boundary (dashed).}
\label{fig:annulus}
\end{figure}

The target function throughout is a radial chirp, $f(x,y)=\sin(x^{2}+y^{2})$, trained on
$1200$ points from the annular wedge $r\in[1,2.2]$, $\theta\in[0,\pi/2]$ and
probed on fresh points in distribution, at unseen angles, and at unseen radii
$r\in[2.2,4]$ (Figure~\ref{fig:annulus}). The first observation is the criterion's central distinction
appearing inside a training run. A sine-activation network
\citep{sitzmann2020siren}, whose basis is
matched to the target, improves training loss over ReLU by one to two orders of
magnitude ($2.5$--$46\times10^{-6}$ against $2.9\times10^{-4}$ across a sweep of
seeds and initialization scales) with no corresponding gain at unseen radii,
error $1.10$--$1.31$ against $2.33$, and the best-fitting configuration is
one of the worst extrapolators. Training loss measures agreement on the region
where agreement was calibrated, which is exactly the quantity that, according to
Criterion~\ref{crit:main}, does not transfer, so a model of this kind has no
internal signal on the difference between fitting the target law and fitting the
training region.

A system that commits to a hypothesis by explicit selection does. Given
only raw coordinates and a fixed operator set, the bottom-up enumerative search
of \S\ref{sec:follows} recovers $\sin(x^{2}+y^{2})$ with coefficient one
and intercept zero, at a selection residual of $8.8\times10^{-33}$, and the
recovery is visible from the inside, on training data alone: the winner
separates from the best semantically distinct runner-up by roughly thirty
orders of magnitude in residual. Removing $\sin(\cdot)$ from the operator set, an
ablation of an operator the answer needs rather than of the answer itself,
leaves the top candidates within $4\%$ of one another at residual
$2.2\times10^{-2}$, returned with no warning: hypothesis wrong, and diagnosably so, though the
diagnosis is not the missing gap. Candidates are scored throughout this section by
a description length, the Bayesian information criterion of
\citet{schwarz1978bic}, $n\log(\mathrm{mse})+k\log n$ for $n$ training points and
a candidate with $k$ parameters, which reads as the length of a two-part code that
transmits the parameters and then the residual; differences in it are quoted in
nats on the deviance scale. A winner is isolated when it leads by more than
$2\log(|\mathcal{H}|-1)$, below which it holds less than half of the posterior
mass under a uniform prior on the class $\mathcal{H}$. The ablated winner is still
isolated, by $64$ nats against a threshold of $27$; what it lacks is adequacy, a
residual four orders of magnitude above the noise floor, and a companion paper \citep{rocha2026certificate} shows why this is the part of the signal that generalizes. The separation is a property
of commitment, not of discreteness, in the same sense as
\S\ref{sec:graded}: a continuous three-parameter family
$A\sin(\omega u+\varphi)$ over $u=x^{2}+y^{2}$, scored by the same
description-length criterion against the second-best mode of its likelihood
surface, earns a certificate on the same scale as the discrete search's, up to
$71.0$ nats per point against the search's $70$, stable under grid refinement,
while a Neural Network blending the same basis functions by unconstrained gradient
descent obtains none of the significant separation obtained in the two previous examples: the discrete search's comparison against its runner-up and the continuous family's comparison against its second-best mode.

\subsection{Distance is not unanswerability}
\label{sec:determined}

The induced symbolic expression entails two global constraints that were supplied by
no one: the range bound $|f|\le1$, and periodicity in $u=x^{2}+y^{2}$ with
period $2\pi$. The training wedge covers $u\in[1,4.84]$, $0.61$ of a period, so
folding a distant query's $u$ back modulo $2\pi$ lands part of the radial
extrapolation region inside the training support (Figure~\ref{fig:annulus}). The
determined fraction is $56.2\%$ rather than $0.61$, because the query range spans
$1.776$ periods and the last folded copy is truncated
(\S\ref{sec:derivations}); the measured count over the queries is $56.3\%$.
Relative to the induced constraint the training data already provides the answer on that subset, and evaluating the induced symbolic expression there, we obtain an error of $1.3\times10^{-16}$. The remaining $43.7\%$ is novel, in the sense that data plus constraint do not determine it.

\begin{table}[htbp]
\centering
\small
\caption{Radial queries by periodicity partition. A $K{=}6$ deep ensemble
fails equally on both subsets and is falsely confident on both (ratio is
disagreement over error); a Nearest-Neighbour distance detector abstains on
both, which on the determined $56.3\%$ is a false abstention. ``N/A'' marks
what training data alone can certify.}
\label{tab:partition}
\begin{tabular}{lcccc}
\toprule
radial subset & ensemble err. & ratio & symbolic err. & distance detector \\
\midrule
determined ($56.3\%$) & $4.43$ & $0.23$ & $1.3\times10^{-16}$ & abstains (falsely) \\
novel ($43.7\%$) & $4.21$ & $0.24$ & N/A & abstains (correctly) \\
\bottomrule
\end{tabular}
\end{table}

Every additive estimator ranks this partition backwards. The determined
queries sit farther from the training data on average, mean
Nearest-Neighbour distance $1.16$ against $0.82$, and draw more
ensemble disagreement, $0.78$ against $0.57$: as predictors for determined queries, the two signals score AUC: $0.69$ and $0.67$ in the wrong
direction, moderate effect sizes pointed backwards rather than chance,
because the periodic bands are placed by the law, to which neither signal has
access. A distance-aware detector therefore abstains precisely on the queries
whose answers the training data already provides. Being far from the training data and
being unanswerable are distinct, measurable conditions, and the quantity that
separates them exists in the hypothesis class.

\subsection{The competence boundary coincides with the inductive bias}
\label{sec:boundary-bias}

The propagation rule of \S\ref{sec:composition} has an epistemic reading, and
it is testable. On a radially symmetric target $g(r)=r^{-2}$, an
$\mathrm{O}(2)$-invariant ensemble, which computes $r$ exactly and fits $g$,
stays accurate and calibrated at unseen angles, where only its exact component
is exercised ($0.049\to0.046$ error, ratio $1.78\to1.85$), and at unseen radii,
where its fitted component is consulted, it is worse than the generic
Neural Network (error $4.88$ against $2.42$) without its disagreement becoming
reliable. The generic ensemble on the same axis is the worst case: error
$2.42$ at ratio $0.09$, consensus without competence, because every ReLU
member extrapolates affinely in the same direction for the same
representational reason, so their agreement certifies nothing
\citep{lakshminarayanan2017ensembles, ovadia2019trust}. The model is exact
precisely where its exact component acts, it fails silently precisely where its
fitted component is consulted, and its self-report has the same boundary as
its competence. Nothing in the uncertainty machinery sees that boundary; it is
visible only from the architecture, by the inspection
\S\ref{sec:composition} describes.

\section{Epistemic Dependence: Who Supplies the Structure}
\label{sec:dependence}

\subsection{The pattern in the successful cases}

There is a common thread in the successful cases. Every architecture that achieves exactness on some region does so because a person
injected, in advance, a representation structurally equivalent to the process
generating the data on that region. Equivariant networks are
exact along the orbits of a group that a designer chose
\citep{cohen2016groupequivariant, bronstein2021geometric, villar2021scalars}. In
Hamiltonian and Lagrangian networks the conservation of energy is exact because a
symplectic structure was written into the architecture, while the Hamiltonian itself is left as an arbitrary fitted Neural Network \citep{greydanus2019hamiltonian,
cranmer2020lagrangian}. ESBN binds variables because indirection was built into its
memory \citep{webb2021esbn}. GraIL generalizes to unseen entities because entity
identity was designed out of its representation \citep{teru2020grail}. In every case, the model did not discover the structure; it was given it, and the model is exact precisely where this structure operates and nowhere else.

This is a coherent and often excellent engineering strategy, and we are not arguing
against it. We are observing what it costs. Structure supplied by a designer is
bounded by what the designer knew to supply, and it does not accumulate: a Neural Network
given the rotation group does not thereby acquire the inverse-square law, and one
given a symplectic form does not thereby acquire the Hamiltonian. Each new domain
requires a new specification, produced by a person, and the model contributes
nothing to producing it.

The consequence for scaling is direct. If the exactness a system exhibits was
injected rather than induced, then scaling the system enlarges the fitted part and
leaves the exact part exactly where it was. This is consistent with the finding of
\citet{deletang2023chomsky} that architecture class, and not training data or training time, predicts OOD generalization, and with \citet{campbell2024binding},
who report that frontier vision-language models still exhibit binding failures. It
also explains why apparent extrapolation successes so often turn out, on inspection, to be a human supplying the linearizing encoding: representability is a property of a problem as encoded, and the encoding is where the knowledge
went in.

\subsection{The objection against hand-designed priors, and why it fails}
\label{sec:declared-prior}

One objection is that a hand-designed language bias sneaks in the answer, where a connectionist model learns just from the data.

It does not, because the connectionist model is underdetermined in precisely the
same way. A Neural Network carrying more parameters than data admits infinitely many
weight settings that fit the training set exactly and disagree everywhere off it,
and capacity is not the binding constraint, as is demonstrated by the fact that
such Neural Networks fit randomly assigned labels perfectly well
\citep{zhang2016understanding}. Nothing in the data selects among those settings.
What selects is the architecture, the imposed symmetry group, the implicit
regularization of the optimizer, the initialization scale and the stopping criterion, which are all priors, and none of them comes from the data.

The two paradigms are therefore not divided by whether they depend on a prior, but
by whether the prior is declared. A vocabulary of primitives, once given, can be read, audited, disputed and replaced; a CNN trained by Stochastic Gradient Descent (SGD) has no less of a prior over the space of hypotheses, but it is distributed across
design choices with no single object to point at, and it is fixed at design time
all the same.

Being hand-designed does not carry the cost it appears to, provided the vocabulary is chosen on principled grounds. The design cost is paid once and amortized across a
class of tasks rather than incurred per task, and this amortization is stronger the more general the designed priors are. On the other hand, learning a vocabulary automatically from experience, whether by library learning or predicate invention, is an alternative answer
to the same question and the two compose naturally, a principled seed vocabulary
being exactly the kind of starting point such methods extend; but a learned
vocabulary is itself fixed relative to the distribution of tasks that produced it,
so it relocates the choice into the selection of that distribution rather than
dissolving it.

\subsection{The residual failure modes, and why they are universal}
\label{sec:graceful}

Systems that induce and execute first-order structure are not immune to OOD failure
either, and two failure modes remain that are not exclusive to them. The first is
search. Even when the hypothesis language contains the correct solution, the search
may not find it: the time allowed for training or search may be too short, the signal
being optimized, a loss or a scoring metric, may not single out the correct
hypothesis, or the computation available may not cover the space. This holds for a
Neural Network trained by SGD as much as for a symbolic search, and
\S\ref{sec:follows} returns to it as the combinatorial problem both paradigms share.

The second is missing knowledge, since no system reasons about an entity it has not
been told about. A symbolic engine needs the entity's base relational properties
declared at inference time, as in
$\text{Adjacent}(\text{Entity}_{\text{new}}, \text{Node}_5)$, and without them the
entity cannot be evaluated against universally quantified rules. A Neural Network
needs the same information in a harder form, as weights, so it cannot be told anything
after training without being retrained. The difference that matters is detectability.
The input space of a Neural Network is total: an unmapped entity, a zero vector and a
freshly initialized embedding all produce confident outputs, so the epistemic failure
is concealed by the geometric one and surfaces as a confident wrong answer, with no
object to point at. A symbolic engine abstains when a required premise is missing,
and the missing premise can be located and supplied.

The cost is coverage. An engine that abstains whenever a premise is missing also
abstains where interpolation would have succeeded, and on a forced-choice benchmark
an abstention and a wrong answer score identically, so the value of abstaining is
realized only in deployment, and only where the surrounding system treats
``unknown'' as an outcome. Leaving search aside, which both paradigms share, DL
suffers both the geometric and the epistemic failure, with the second hidden behind
the first, while systems that execute induced structure suffer only the epistemic
one, and it is visible.

\section{What Follows}
\label{sec:follows}

If our criterion (Criterion~\ref{crit:main}) is right, the requirement on a system aiming at OOD generalization is to induce a structure that can be executed exactly. Two mature symbolic induction paradigms do this over restricted languages.

\paragraph{Symbolic Regression, for the continuous face.}

In continuous spaces, Neural Networks fail because their piecewise-linear topology traps
them in rigid asymptotes \citep{hein2019relu, xu2020how}. SR
abandons continuous optimization: instead of approximating a function by tuning
weights to place bounding hyperplanes, it searches a discrete combinatorial space
of mathematical expressions for the analytical form that generated the data
\citep{schmidt2009distilling}. Genetic programming \citep{koza1992genetic}, sparse
regression \citep{brunton2016sindy}, and modern systems combining the two
\citep{udrescu2020aifeynman, cranmer2023pysr} build and mutate abstract syntax
trees over algebraic operators, lifting the empirical relationship into its
algebraic structure. Once $f(x) = x^2$ is recovered as a formula rather than as a
graph, extrapolation is analytically exact and indifferent to input magnitude along every dimension. The discrete space of expressions is very large, and these methods are ways of searching it, each with limitations of its own, which we take up together with those of ILP below.

\paragraph{Inductive Logic Programming, for the relational face.}

Structured relational domains require the parallel shift. ILP searches a space of
clauses and returns a program executed by resolution \citep{muggleton1991inductive,
cropper2020inductive}. A clause with variables can be applied to constants never seen, because nothing in it is referring to a constant, which is exactly the property the
embedding models of \S\ref{sec:relational-face} lack.

\paragraph{The combinatorial space, for both.} The standing objection to both paradigms is the combinatorial one. In SR the space of expressions grows exponentially with expression size, and finding the best expression is NP-hard \citep{virgolin2022srnphard}; genetic programming, sparse regression and their combinations search it heuristically, and benchmark comparisons find that they still fail to recover a substantial fraction of known ground-truth equations \citep{lacava2021srbench}. In ILP, hybrid search unifies bottom-up clause construction, which
bounds the space using empirical evidence, with top-down refinement, which
maintains logical consistency, pruning the subsumption lattice from both ends.
Declarative language biases such as meta-interpretive learning
\citep{muggleton2015metainterpretive} constrain the syntax of the hypothesis space
to structurally sound templates. And compiling the induction problem into satisfiability or Answer Set Programming (ASP), as in ILASP \citep{law2014answersetprograms} and Popper \citep{cropper2021popper}, brings conflict-driven clause learning to traverse the search space more efficiently. These methods move the boundary without removing it: the hypothesis space still grows combinatorially with program length, and learning recursive programs is harder still \citep{cohen1995recursive, cropper2020inductive}. The size of the combinatorial space is the main weakness the two paradigms share, and it is not yet solved.

\subsection{A note on sample efficiency, and on not quantifying it}
\label{sec:sample-efficiency}

One practical argument often made for symbolic induction is that it reaches a given level of confidence from far less data than a Neural Network, and at the level
of mechanism the argument is sound. Constraining a hypothesis space to a finite,
syntactically bounded family of discrete structures trades expressive flexibility
for statistical efficiency, which is the classical Occam-style observation. The
empirical record bears it out, with ILP systems routinely inducing correct programs
from tens or hundreds of examples where a connectionist model of comparable nominal
expressive power requires orders of magnitude more \citep{muggleton1991inductive,
muggleton1995inverse}.

We state that advantage qualitatively rather than numerically. Setting a Vapnik-Chervonenkis (VC) bound
for a piecewise-linear network against a finite-class bound for a bounded clause
space compares quantities that are not commensurable, and the mismatch runs in
opposite directions. The VC bound is agnostic and worst case, and is loose to the
point of vacuity for Neural Networks that in practice generalize far better than it
predicts, which inflates the Neural Network's figure. The finite-class bound presupposes
that the target clause already lies inside the hypothesis space, which is precisely
the difficult part, and is silent on whether search finds it, which deflates the
symbolic figure. A ratio assembled from the two would be an artefact of those
choices rather than a measurement. We therefore report the mechanism and the empirical record, and decline to quantify it.

What a symbolic engine saves in data it spends in search: it exchanges a statistical problem for a combinatorial one, as we have seen previously. Neither paradigm escapes a cost, and which is cheaper depends on the domain and is not a general fact.

\subsection{Program synthesis as the general form}

The two symbolic induction paradigms described are not sufficient for the general case. Analytical expressions lack state and iteration; logical clauses encode procedures awkwardly; and many generating mechanisms of interest correspond to procedural algorithms. Program synthesis is the general form, and its costs
are real: undecidability, termination, verification, and a worse search problem than
either restriction. The mitigations are the familiar ones, a Domain-Specific Language (DSL) as language bias, learned guidance for the search, and library learning to
accumulate reusable structure \citep{ellis2021dreamcoder}. Our own work applies ILP to the Abstraction and Reasoning Corpus in this spirit
\citep{rocha2024programsynthesis}, and earlier work on Relational Reinforcement Learning found the same pattern from another direction, improving over DL baselines while supplying domain knowledge \citep{rocha2020towardsagi,
rocha2020overcomingrl}.

\subsection{Placing each bias where it is required}
\label{sec:heterogeneous}

The argument so far has treated the hypothesis language as a single choice applied
uniformly. This assumption limits the most general case. Within one task some sub-relations are relational, concerning which entity stands in
which relation to which and under what condition; some are numeric, an offset, a length, a scale factor; and some are procedural, in that the output
of one stage is the input of the next. Each has its own natural language.

The choice among them is not only a matter of convenience or of search cost, and
this is where the criterion re-enters. The language is the representation the
system will execute when a prediction is made, so forcing a relational or a
procedural sub-relation into a continuous piecewise-affine surrogate does not
merely make it expensive to express, it makes it impossible to express exactly:
such a surrogate is not structurally equivalent to a quantified rule, which must
hold of constants it was never fitted on, nor to an ordered succession of states,
which has no representation at all in a map from inputs to outputs. Choosing the
language is therefore choosing which sub-relations the system can compute exactly rather than approximate, and a sub-relation placed in the wrong language is outside our criterion (Criterion~\ref{crit:main}) before any learning begins.

Forcing a single language to carry all three is paid for in description length, and
description length is the exponent of the search. A numeric dependency such as ``the displacement equals the entity's width less one'' occupies a single literal when the
vocabulary contains a bounded algebraic form, and otherwise requires either a chain
of literals or a distinct constant for every value the quantity may take. A
multi-stage transformation occupies one short clause per stage when the search may
pose each stage as its own problem, and otherwise becomes a single clause whose length is the sum of the stages', or a recursive clause, and learning recursive clauses is considerably more expensive \citep{cohen1995recursive}. Stated as complexity, if a task decomposes into $S$ stages each solvable at length $L$ over a vocabulary of size $V$, a monolithic search is $\mathcal{O}(V^{SL})$ where a staged search is $\mathcal{O}(S \cdot V^{L})$. When the stages are instances of one step applied repeatedly, a recursive program of a base clause and a recursive clause, each of length at most $L$, can replace them, so the candidates number $\mathcal{O}(V^{2L})$; but each candidate must be unfolded to depth $S$ to be checked, and since termination of a candidate recursive program is undecidable in general, the search must also impose a depth bound, which gives $\mathcal{O}(S \cdot V^{2L})$, a factor $V^{L}$ above the staged search and available only when the stages are homogeneous. The reduction from monolithic to staged search is not a constant factor, and it is the difference between a search that
terminates within a budget and one that does not.

Two of the three languages carry a further requirement that is not merely a matter of
vocabulary size.

Take the relational part first. Its natural language is a logic programming one, and the reason is Kowalski's decomposition of a
program into logic and control \citep{kowalski1979algorithm}. A synthesizer coming up with relational programs in an imperative or functional language must produce both halves: the declarative
content of the target relation, and an execution plan that iterates over entities,
binds variables, orders tests, routes branches and unwinds state when a partial
construction fails. The search space is the product of the two, and the second factor
carries no information about the task. A logic programming language supplies the
control half once, as the fixed semantics of resolution, unification and
backtracking, so the synthesized object is the logic alone. Two consequences follow.
Constraints that hold simultaneously are expressed by conjunction rather than
by sequencing: where a functional composition $f(g(x))$ imposes an order the problem
does not impose, a conjunction of literals asks only for a binding satisfying all of
them at once. And context dependence is expressed by multiple clauses for the same head rather than by explicit branching, so contexts add to the search space
rather than multiplying it, and a clause whose guards are unmet simply fails and
yields to the next.

Take next the procedural part, where the requirement is sharper, because it is a limitation of FOL itself rather than of a particular programming language. Pure FOL
is atemporal: it states what holds, not what happens next, and has no native notion of
a state that is superseded. A transformation whose stages must run in order cannot be
stated as an ordinary clause. It can be forced into one, by reifying time and
quantifying over situations in the manner of the Situation Calculus
\citep{blackburn2001situation}, or by resorting to extra-logical database side effects,
and both routes lengthen the clause with machinery encoding the passage of state
rather than the content of the transformation. The alternative is to leave the logic
atemporal and place the sequencing outside it, in a procedure that applies an induced
program, observes the resulting state, and poses the next stage as a fresh induction problem against that state, which is exactly what we did with our proposed system, ILPAR \citep{rocha2024programsynthesis}. This converts the exponent above into a coefficient, and
it has an information-theoretic reading. Universal induction identifies the best
explanation of the data with the shortest program that generates it
\citep{chaitin1987ait}, and searching directly for that program is intractable. The
theory of incremental compression \citep{franz2019william} observes that the tractable
form of the same objective is a succession of partial compressions, each removing some
residual structure and handing the remainder to the next, and a procedural loop over
atemporal inductions is that construction carried out in FOL. Domains with this character are not rare: physical simulation, visual reasoning benchmarks built from staged spatial edits \citep{chollet2019measure}, and matrix-completion tests of fluid intelligence such
as Raven's Progressive Matrices \citep{carpenter1990ravens} all present transformations
whose stages are ordered and whose intermediate states are never observed.

Three costs attach and should not be glossed over. Something must perform the
decomposition, and that something is itself an inductive bias, so an incorrect
decomposition is not repaired by correct per-part biases downstream of it. The parts
must interoperate, since the output of one paradigm has to be a legal input to another.
And composing languages enlarges the vocabulary, so the gain depends on the program length $L$ shrinking faster than the program vocabulary $V$ grows, which is an empirical claim about a domain and not a theorem.

What this analysis yields is a specification of four parts rather than a system.

\begin{enumerate}
\item A \textbf{relational core expressed in a logic programming language},
supplying the universal quantification and dynamic variable binding that \S\ref{sec:relational-face} showed the connectionist paradigm lacks, in a form where the synthesized object is the logic and the control is baked into the interpreter.
\item A \textbf{bounded algebraic primitive} for numeric dependencies, so a
quantitative relation between a target quantity and a property of the situation costs
one literal rather than an enumeration of constants.
\item An explicit \textbf{procedural stage structure} external to the logic, so state
transformation is carried by sequence rather than clause length, which is what converts
the exponent into a coefficient and makes incremental compression available in a FOL setting.
\item A \textbf{vocabulary chosen on principled grounds} rather than for convenience,
since it is the vocabulary that renders the induction well posed at all.
\end{enumerate}

One further element is permitted by this specification, because our criterion (Criterion~\ref{crit:main}) is what licenses it. A learned model may propose or rank candidate fragments
during search without appearing anywhere in the inference path of the induced
hypothesis, so search guidance is a legitimate place for a neural component in a way that inference is not, as explained in \S\ref{sec:composition}. The extrapolation guarantee is a property of what is executed,
and a Neural Network that only influences which candidates are examined leaves that property
untouched.

\subsection{The evidence from abstraction benchmarks}
\label{sec:arc}

The evidence from ARC-AGI, the benchmark built to measure exactly the OOD generalization capacity
\citep{chollet2019measure}, is relevant here, because the distinction that organizes it is the one this paper is about. ARC-AGI state-of-the-art approaches divide into induction, which infers a latent function or relation and then applies it, and transduction, which predicts the test output directly without ever forming a reusable rule
\citep{li2024induction}. A transductive system consults a fitted model at
inference by construction: whatever it has learned is never separated from the act of
predicting, so there is nothing that could be executed exactly.

Transductive systems have led the reported private-set results \citep{chollet2026arc},
and we address the strongest statement of the case for them directly. \citet{cole2025babybathwater} argue that ``fully committing to deep
learning's capacity to acquire novel abstractions yields state-of-the-art performance on
ARC,'' and their methodological claim is the interesting part: that both ``the neural network and the optimizer (rather than just a pre-trained network)'' should be
treated ``as integral components of the inference process.'' Test-time fine-tuning is
therefore not the case \S\ref{sec:dependence} argues against. It is neither a human
injecting structure in advance nor a frozen approximation consulted blindly, but a fresh
approximation fitted per task, which is a third position our framing must accommodate
rather than dismiss. What it does not do is produce anything that could be executed
exactly: the artefact it fits is a set of weights, consulted at inference time, so it falls on
the approximate side of Criterion \ref{crit:main} however well it scores.

The comparison that bears directly on our criterion has been run.
\citet{li2024induction} train inductive and transductive models on the same problems with
the same architecture, differing only in whether a Python program is produced and
executed or an output is predicted, and find that the two ``solve different kinds of test
problems.'' Inductive program synthesis ``excels at precise computations, and at composing
multiple concepts,'' while transduction ``succeeds on fuzzier perceptual concepts.'' This is our criterion's prediction, arrived at independently: exact execution is what buys
precise computation and composition, since a rule composes with another rule only if each
computes what it says, whereas an approximation composed with an approximation degrades.
Their finding that ensembling the two approaches human-level performance is equally
consistent with the position taken here, and with the specification of
\S\ref{sec:heterogeneous}: the two do different jobs and a system may want both, placed
where each is appropriate.

We therefore make no claim that symbolic induction alone is empirically sufficient, and we
attach no weight to any particular standing, which is a moving quantity and a poor
foundation for an argument about representation. The claim is that where a system generalizes to a novel regime something in it computed the mechanism exactly,
and that the open question is whether that something was induced by the system or supplied
by a person. The prediction that follows is about problem classes rather than totals: the
transductive share should concentrate where approximation suffices, and a composition
boundary should appear where it does not, which is what \citet{li2024induction} report and
what further work of that design could test directly. The defining theme of the ARC-AGI 2025 technical report \citep{chollet2026arc} is the refinement loop, ``a per-task iterative program optimization loop guided by a feedback signal'', with the strongest instances evolving program solutions in Python. A per-task search for a program that is then executed is symbolic induction, whatever the search is
made of.

\section{Scope, Limitations, and Refutation Conditions}
\label{sec:limits}

Our criterion is not a theorem. We have not proved that exactness is
necessary for OOD generalization; we have argued that it predicts the
observed pattern of successes and failures better than Neural Network architecture, capacity, data, or scale, and that it sorts neuro-symbolic systems in a way their surface taxonomy \citep{kautz2022third} does not.

The unification of the relational and continuous faces is a conceptual claim, and its parts are established results due to the work of others. The continuous analysis is scoped to unbounded-output
regression networks, for the reasons in \S\ref{sec:scope}. The relational analysis is scoped to DL architectures representing entities as free per-entity parameters, and is false
outside that scope, as \S\ref{sec:right-scope} states. The complexity statement in
\S\ref{sec:heterogeneous} assumes a bias decomposition is available and correct, which is
itself a bias and not a given.

What would refute the position. A fully differentiable system that consults a fitted
approximation at inference and nonetheless exhibits exact, non-degrading agreement with a
generating mechanism on shifts chosen adversarially, in the case where this mechanism is not a piecewise-linear function, which is the form a piecewise-affine Neural Network represents exactly. Or a demonstration that architectural
exactness accumulates across domains without per-domain human specification, which would
undercut \S\ref{sec:dependence}. Or a scaling result in which OOD
generalization on a mechanism outside the representable class improves with data or
parameters at a rate that does not vanish, which would contradict the representational
reading directly.

\section{Conclusion}

Deep networks do not always fail at extrapolation. They fail at extrapolating mechanisms they cannot represent exactly, which is the most common case, but they succeed, indefinitely far from any data, on the
mechanisms they can. The same statement covers the relational case once it is scoped
correctly: an architecture whose representation of an entity is a fitted parameter cannot
apply a quantified rule to an entity it never fitted, and an architecture that represents
bindings can. What unites these is not a shared pathology but a shared boundary, the
boundary of exact representability, and what matters at inference is that the operation
performed computes the intended relation rather than approximating it where the data
happened to be.

That criterion is more permissive than the usual symbolic prescription and more demanding
than the usual neuro-symbolic one. It permits continuous arithmetic, graded semantics, and
approximate search. It refuses only the consultation of a fitted continuous approximation at the moment an inference is made. And it settles more than accuracy: the boundary of what a model can
know about its own predictions is the same boundary, which is why abstention cannot be
estimated onto a hypothesis class that cannot state what is being abstained from.

The uncomfortable truth for Machine Learning is that most of the exactness in contemporary systems was put
there by people. The common repairs, from hardcoded symmetries to external memory to
symplectic structure, reach exactness only because a person injected a representation
structurally equivalent to the process that generated the data. That is not an argument
against building it in. It is an argument for the harder project of inducing it, because
injected structure is bounded by what its designer anticipated and induced structure is not.
Passing that bound requires architectures able to induce exact representations
autonomously, rather than fitting continuous surrogates constrained to representations
such as the piecewise-affine maps of MLPs, which are incorrect unless the generating
mechanism is itself of that form, and it rarely is. Such surrogates approximate within the training distribution. Their residual error rarely vanishes on the support, and even where a sufficiently flexible approximant drives it to the arithmetic floor there, as a high-degree polynomial can, it diverges outside the support and compounds under composition.

\appendix

\section{Mathematical derivations}
\label{sec:derivations}

Four results are stated in the body without their algebra, to not interrupt the argument. They are collected here so that a reader who wants
to check them does not have to reconstruct them.

\subsection{The chord residual and the factor of two (\S\ref{sec:datahungry})}

Take $f(x) = x^{2}$ and an interval $[a,b]$ of width $h = b - a$. The affine
interpolant through the endpoints is $L(x) = (a+b)x - ab$, and
\[
L(x) - x^{2} = -x^{2} + (a+b)x - ab = (x-a)(b-x),
\]
which is zero exactly at the two nodes, positive strictly between them, and
maximal at the midpoint, where it equals $(h/2)^{2} = h^{2}/4$. Since $f'' = 2$,
that is $h^{2}f''/8$, the bound quoted in \S\ref{sec:datahungry}, and it is
attained rather than merely bounded. The residual vanishes only at the nodes, so
interpolating the samples exactly leaves it everywhere else on the interval.

The best affine approximant on $[a,b]$ is not the chord. By the Chebyshev
equioscillation theorem the best uniform approximant of degree one is
characterized by an error that equioscillates at three points, which here is
$L$ displaced downward by half the maximum chord deviation: the displaced map
errs by $h^{2}/8$ below the chord's nodes and $h^{2}/8$ above it at the midpoint,
with equal magnitude and alternating sign at $a$, $(a+b)/2$ and $b$. Its maximum
error is therefore $h^{2}/8 = h^{2}f''/16$, half the chord's. For a general
twice-differentiable $f$ the same two statements hold with $\max|f''|$ in place
of $f''$, giving $h^{2}\max|f''|/8$ and $h^{2}\max|f''|/16$. Panel (c) of
Figure~\ref{fig:residual} plots both curves against the measured residuals.

\subsection{Two GELU units of opposite input-weight sign (\S\ref{sec:activations})}

Write $\Phi$ for the standard normal cumulative distribution function and
$\mathrm{erf}$ for the error function,
$\mathrm{erf}(z) = \frac{2}{\sqrt{\pi}}\int_{0}^{z} e^{-t^{2}}\,\mathrm{d}t$, so that
$\Phi(z) = \tfrac{1}{2}\left(1 + \mathrm{erf}(z/\sqrt{2})\right)$. A GELU unit is
$\mathrm{GELU}(x) = x\Phi(x)$. For the normalized pair, two units with input
weights $+1$ and $-1$ and output weights both $1$,
\[
\mathrm{GELU}(x) + \mathrm{GELU}(-x) = x\Phi(x) - x\Phi(-x)
= x\left(2\Phi(x) - 1\right) = x\,\mathrm{erf}\!\left(x/\sqrt{2}\right),
\]
using $\Phi(-z) = 1 - \Phi(z)$ at the second step and the definition of
$\mathrm{erf}$ at the third. Since $\mathrm{erf}(z) \to \pm 1$ as
$z \to \pm\infty$, the sum is asymptotic to $|x|$, so it grows without bound on
both tails although each unit is asymptotically zero on one of them. A general
pair $a_{1}\mathrm{GELU}(w_{1}x)$ and $a_{2}\mathrm{GELU}(w_{2}x)$ with
$w_{2} = -w_{1}$ scales the two tails separately, with asymptotic slopes
$a_{1}w_{1}$ and $a_{2}w_{2}$, which is the decomposition measured in panel (c)
of Figure~\ref{fig:asymptotics}.

\subsection{The determined fraction on the radial chirp (\S\ref{sec:determined})}

Training covers $r \in [1, 2.2]$, so $u = x^{2}+y^{2}$ ranges over $[1, 4.84]$;
the radial query region is $r \in [2.2, 4]$, so $u$ ranges over $[4.84, 16]$. The
recovered expression is $2\pi$-periodic in $u$, so a query at $u$ is determined by
the training data exactly when $u \bmod 2\pi$ falls in the training interval.

The training interval has length $3.84$, which is $0.611$ of a period, but that
is not the determined fraction, because the query interval has length $11.16$,
which is $1.776$ periods and not a whole number of them, so the folded copies of
the training interval do not tile it. Two copies meet the query interval:
$[1+2\pi,\, 4.84+2\pi] = [7.283,\, 11.123]$, which lies inside it and contributes
its full $3.84$, and $[1+4\pi,\, 4.84+4\pi] = [13.566,\, 17.406]$, which is
truncated at $u = 16$ and contributes $2.434$. The determined length is $6.274$
out of $11.16$, or $56.2\%$.

Points drawn uniformly by area give $u$ uniformly, since $u = r^{2}$ gives
$\mathrm{d}A = r\,\mathrm{d}r\,\mathrm{d}\theta = \tfrac{1}{2}\,\mathrm{d}u\,\mathrm{d}\theta$,
which does not depend on $r$. The geometric fraction is therefore what a finite
sample estimates, and the count over the sampled queries is $56.3\%$. The
binomial standard deviation of such an estimate is
$\sqrt{0.562 \cdot 0.438 / n}$, about one point at $n = 2000$, so the two
numbers agree.

\subsection{The propagation bound (\S\ref{sec:composition})}

Let $g_{1},\dots,g_{n}$ be the exact stages of a pipeline and
$\hat g_{1},\dots,\hat g_{n}$ the fitted stages standing in for them. Two
quantities describe stage $k$: a uniform error $\varepsilon_{k}\ge 0$ and a
Lipschitz constant $L_{k}\ge 0$ for the exact stage, combined in the single
hypothesis
\[
\bigl|\hat g_{k}(u)-g_{k}(v)\bigr| \;\le\; \varepsilon_{k}+L_{k}\,|u-v|
\qquad\text{for all } u,v,
\]
which says that the fitted stage is wrong by at most $\varepsilon_{k}$ on an
input the exact stage would have received, and that the exact stage magnifies a
discrepancy it receives by at most $L_{k}$. Write $x_{k}=g_{k}(x_{k-1})$ and
$\hat x_{k}=\hat g_{k}(\hat x_{k-1})$ from a common input $x_{0}=\hat x_{0}$,
and let $e_{k}=|\hat x_{k}-x_{k}|$. The hypothesis applied with
$u=\hat x_{k-1}$ and $v=x_{k-1}$ gives the recursion
\[
e_{0}=0,\qquad e_{k}\;\le\;\varepsilon_{k}+L_{k}\,e_{k-1},
\]
and unrolling it by induction on $k$ gives
\[
e_{n}\;\le\;\sum_{k\le n}\varepsilon_{k}\prod_{j>k}L_{j},
\]
since the inductive step
$e_{k}\le\varepsilon_{k}+L_{k}\sum_{i<k}\varepsilon_{i}\prod_{i<j<k}L_{j}
=\sum_{i\le k}\varepsilon_{i}\prod_{i<j\le k}L_{j}$ absorbs the new factor
$L_{k}$ into every earlier product. Three consequences follow directly. If every
$\varepsilon_{k}$ is zero the bound is zero whatever the constants, which is the
statement that exactness composes and needs no Lipschitz hypothesis at all; the
bound is monotone in each $\varepsilon_{k}$, so a stage that contributes no
error contributes nothing to the composite; and if every downstream constant is
at least $c$, the bound grows by at least a factor of $c$ per stage, which is the
compounding claim in the body. Nothing in the argument uses continuity,
differentiability or dimension, and $|u-v|$ may be read as any metric.

\bibliographystyle{plainnat}
\bibliography{references}

\begin{thebibliography}{93}
\providecommand{\natexlab}[1]{#1}
\providecommand{\url}[1]{\texttt{#1}}
\expandafter\ifx\csname urlstyle\endcsname\relax
  \providecommand{\doi}[1]{doi: #1}\else
  \providecommand{\doi}{doi: \begingroup \urlstyle{rm}\Url}\fi

\bibitem[Abiteboul et~al.(1995)Abiteboul, Hull, and
  Vianu]{abiteboul1995foundations}
Serge Abiteboul, Richard Hull, and Victor Vianu.
\newblock \emph{Foundations of Databases}.
\newblock Addison-Wesley, 1995.
\newblock ISBN 0201537710.

\bibitem[Ahmed et~al.(2022)Ahmed, Teso, Chang, Van~den Broeck, and
  Vergari]{ahmed2022semantic}
Kareem Ahmed, Stefano Teso, Kai-Wei Chang, Guy Van~den Broeck, and Antonio
  Vergari.
\newblock Semantic probabilistic layers for neuro-symbolic learning.
\newblock In \emph{Advances in Neural Information Processing Systems
  (NeurIPS)}, 2022.

\bibitem[Ahuja et~al.(2021)Ahuja, Caballero, Zhang, Gagnon-Audet, Bengio,
  Mitliagkas, and Rish]{ahuja2021invariance}
Kartik Ahuja, Ethan Caballero, Dinghuai Zhang, Jean-Christophe Gagnon-Audet,
  Yoshua Bengio, Ioannis Mitliagkas, and Irina Rish.
\newblock Invariance principle meets information bottleneck for
  out-of-distribution generalization, 2021.
\newblock URL \url{https://arxiv.org/abs/2106.06607}.

\bibitem[Azulay and Weiss(2019)]{azulay2019transformations}
Aharon Azulay and Yair Weiss.
\newblock Why do deep convolutional networks generalize so poorly to small
  image transformations?
\newblock \emph{Journal of Machine Learning Research}, 20\penalty0
  (184):\penalty0 1--25, 2019.

\bibitem[Badreddine et~al.(2022)Badreddine, d'Avila Garcez, Serafini, and
  Spranger]{badreddine2022logictensor}
Samy Badreddine, Artur d'Avila Garcez, Luciano Serafini, and Michael Spranger.
\newblock Logic tensor networks.
\newblock \emph{Artificial Intelligence}, 303:\penalty0 103649, 2022.
\newblock \doi{10.1016/j.artint.2021.103649}.

\bibitem[Barcel{\'o} et~al.(2020)Barcel{\'o}, Kostylev, Monet, P{\'e}rez,
  Reutter, and Silva]{barcelo2020logical}
Pablo Barcel{\'o}, Egor~V. Kostylev, Mikael Monet, Jorge P{\'e}rez, Juan
  Reutter, and Juan-Pablo Silva.
\newblock The logical expressiveness of graph neural networks.
\newblock In \emph{International Conference on Learning Representations
  (ICLR)}, 2020.

\bibitem[Bellman(1957)]{bellman1957dynamic}
Richard Bellman.
\newblock \emph{Dynamic Programming}.
\newblock Princeton University Press, Princeton, NJ, 1957.

\bibitem[Bender and Koller(2020)]{bender2020climbing}
Emily~M. Bender and Alexander Koller.
\newblock Climbing towards {NLU}: On meaning, form, and understanding in the
  age of data.
\newblock In \emph{Proceedings of the 58th Annual Meeting of the Association
  for Computational Linguistics (ACL)}, pages 5185--5198, 2020.
\newblock \doi{10.18653/v1/2020.acl-main.463}.

\bibitem[Blackburn et~al.(2001)Blackburn, Kamps, and
  Marx]{blackburn2001situation}
Patrick Blackburn, Jaap Kamps, and Maarten Marx.
\newblock Situation calculus as hybrid logic: First steps.
\newblock \emph{Lecture Notes in Computer Science}, pages 253--260, 2001.

\bibitem[Bordes et~al.(2013)Bordes, Usunier, Garc\'{i}a-Dur\'{a}n, Weston, and
  Yakhnenko]{bordes2013transe}
Antoine Bordes, Nicolas Usunier, Alberto Garc\'{i}a-Dur\'{a}n, Jason Weston,
  and Oksana Yakhnenko.
\newblock Translating embeddings for modeling multi-relational data.
\newblock In \emph{Advances in Neural Information Processing Systems 26
  (NeurIPS)}, pages 2787--2795, 2013.

\bibitem[Bronstein et~al.(2021)Bronstein, Bruna, Cohen, and
  Veli\v{c}kovi\'{c}]{bronstein2021geometric}
Michael~M. Bronstein, Joan Bruna, Taco Cohen, and Petar Veli\v{c}kovi\'{c}.
\newblock Geometric deep learning: Grids, groups, graphs, geodesics, and
  gauges, 2021.
\newblock URL \url{https://arxiv.org/abs/2104.13478}.

\bibitem[Brunton et~al.(2016)Brunton, Proctor, and Kutz]{brunton2016sindy}
Steven~L. Brunton, Joshua~L. Proctor, and J.~Nathan Kutz.
\newblock Discovering governing equations from data by sparse identification of
  nonlinear dynamical systems.
\newblock \emph{Proceedings of the National Academy of Sciences}, 113\penalty0
  (15):\penalty0 3932--3937, 2016.
\newblock \doi{10.1073/pnas.1517384113}.

\bibitem[Campbell et~al.(2024)Campbell, Rane, Giallanza, De~Sabbata, Ghods,
  Joshi, Ku, Frankland, Griffiths, Cohen, and Webb]{campbell2024binding}
Declan Campbell, Sunayana Rane, Tyler Giallanza, Nicol{\`o} De~Sabbata, Kia
  Ghods, Amogh Joshi, Alexander Ku, Steven~M. Frankland, Thomas~L. Griffiths,
  Jonathan~D. Cohen, and Taylor~W. Webb.
\newblock Understanding the limits of vision language models through the lens
  of the binding problem.
\newblock In \emph{Advances in Neural Information Processing Systems
  (NeurIPS)}, 2024.
\newblock arXiv:2411.00238.

\bibitem[Carpenter et~al.(1990)Carpenter, Just, and Shell]{carpenter1990ravens}
Patricia~A. Carpenter, Marcel~A. Just, and Peter Shell.
\newblock What one intelligence test measures: A theoretical account of the
  processing in the {R}aven progressive matrices test.
\newblock \emph{Psychological Review}, 97\penalty0 (3):\penalty0 404--431,
  1990.

\bibitem[Chaitin(1987)]{chaitin1987ait}
Gregory~J. Chaitin.
\newblock Algorithmic information theory.
\newblock In \emph{Information, Randomness and Incompleteness}, pages 33--37.
  World Scientific, 1987.

\bibitem[Chollet(2019)]{chollet2019measure}
Fran\c{c}ois Chollet.
\newblock On the measure of intelligence, 2019.
\newblock URL \url{https://arxiv.org/abs/1911.01547}.

\bibitem[Chollet et~al.(2026)Chollet, Knoop, Kamradt, and
  Landers]{chollet2026arc}
Fran\c{c}ois Chollet, Mike Knoop, Gregory Kamradt, and Bryan Landers.
\newblock Arc prize 2025: Technical report, 2026.
\newblock URL \url{https://arxiv.org/abs/2601.10904}.

\bibitem[Cohen and Welling(2016)]{cohen2016groupequivariant}
Taco Cohen and Max Welling.
\newblock Group equivariant convolutional networks.
\newblock In \emph{Proceedings of the 33rd International Conference on Machine
  Learning (ICML)}, pages 2990--2999, 2016.

\bibitem[Cohen et~al.(2019)Cohen, Weiler, Kicanaoglu, and
  Welling]{cohen2019gauge}
Taco~S. Cohen, Maurice Weiler, Berkay Kicanaoglu, and Max Welling.
\newblock Gauge equivariant convolutional networks and the icosahedral {CNN}.
\newblock In \emph{Proceedings of the 36th International Conference on Machine
  Learning (ICML)}, 2019.

\bibitem[Cohen(1995)]{cohen1995recursive}
William~W. Cohen.
\newblock Pac-learning recursive logic programs: Negative results.
\newblock \emph{Journal of Artificial Intelligence Research}, 2:\penalty0
  541--573, 1995.
\newblock \doi{10.1613/jair.1917}.

\bibitem[Cole and Osman(2025)]{cole2025babybathwater}
Jack Cole and Mohamed Osman.
\newblock Don't throw the baby out with the bathwater: How and why deep
  learning for {ARC}, 2025.
\newblock URL \url{https://arxiv.org/abs/2506.14276}.

\bibitem[Cranmer(2023)]{cranmer2023pysr}
Miles Cranmer.
\newblock Interpretable machine learning for science with {PySR} and
  {SymbolicRegression.jl}.
\newblock \emph{arXiv preprint arXiv:2305.01582}, 2023.
\newblock URL \url{https://arxiv.org/abs/2305.01582}.

\bibitem[Cranmer et~al.(2020)Cranmer, Greydanus, Hoyer, Battaglia, Spergel, and
  Ho]{cranmer2020lagrangian}
Miles Cranmer, Sam Greydanus, Stephan Hoyer, Peter Battaglia, David Spergel,
  and Shirley Ho.
\newblock Lagrangian neural networks.
\newblock \emph{ICLR 2020 Deep Differential Equations Workshop}, 2020.
\newblock URL \url{https://arxiv.org/abs/2003.04630}.

\bibitem[Cropper and Duman\v{c}i\'{c}(2020)]{cropper2020inductive}
Andrew Cropper and Sebastijan Duman\v{c}i\'{c}.
\newblock Inductive logic programming at 30: A new introduction, 2020.
\newblock URL \url{https://arxiv.org/abs/2008.07912}.

\bibitem[Cropper and Morel(2021)]{cropper2021popper}
Andrew Cropper and Rolf Morel.
\newblock Learning programs by learning from failures.
\newblock \emph{Machine Learning}, 110\penalty0 (4):\penalty0 801--856, 2021.
\newblock \doi{10.1007/s10994-020-05934-z}.

\bibitem[Cybenko(1989)]{cybenko1989approximation}
George Cybenko.
\newblock Approximation by superpositions of a sigmoidal function.
\newblock \emph{Mathematics of Control, Signals and Systems}, 2\penalty0
  (4):\penalty0 303--314, 1989.
\newblock \doi{10.1007/BF02551274}.

\bibitem[d'Avila Garcez and Lamb(2023)]{garcez2023neurosymbolic}
Artur d'Avila Garcez and Luis~C. Lamb.
\newblock Neurosymbolic {AI}: the 3rd wave.
\newblock \emph{Artificial Intelligence Review}, 56\penalty0 (11):\penalty0
  12387--12406, 2023.
\newblock \doi{10.1007/s10462-023-10448-w}.

\bibitem[De~Raedt et~al.(2007)De~Raedt, Kimmig, and
  Toivonen]{deraedt2007problog}
Luc De~Raedt, Angelika Kimmig, and Hannu Toivonen.
\newblock {ProbLog}: A probabilistic {Prolog} and its application in link
  discovery.
\newblock In \emph{Proceedings of the 20th International Joint Conference on
  Artificial Intelligence (IJCAI)}, pages 2462--2467, 2007.

\bibitem[Dehghani et~al.(2019)Dehghani, Gouws, Vinyals, Uszkoreit, and
  Kaiser]{dehghani2019universal}
Mostafa Dehghani, Stephan Gouws, Oriol Vinyals, Jakob Uszkoreit, and Lukasz
  Kaiser.
\newblock Universal transformers.
\newblock In \emph{International Conference on Learning Representations
  (ICLR)}, 2019.

\bibitem[Del{\'e}tang et~al.(2023)Del{\'e}tang, Ruoss, Grau-Moya, Genewein,
  Wenliang, Catt, Cundy, Hutter, Legg, Veness, and Ortega]{deletang2023chomsky}
Gr{\'e}goire Del{\'e}tang, Anian Ruoss, Jordi Grau-Moya, Tim Genewein, Li~Kevin
  Wenliang, Elliot Catt, Chris Cundy, Marcus Hutter, Shane Legg, Joel Veness,
  and Pedro~A. Ortega.
\newblock Neural networks and the {C}homsky hierarchy.
\newblock In \emph{International Conference on Learning Representations
  (ICLR)}, 2023.
\newblock URL \url{https://arxiv.org/abs/2207.02098}.
\newblock arXiv:2207.02098.

\bibitem[Deng et~al.(2022)Deng, Ren, Zhang, and Zhang]{deng2022representation}
Huiqi Deng, Qihan Ren, Hao Zhang, and Quanshi Zhang.
\newblock Discovering and explaining the representation bottleneck of {DNN}s.
\newblock In \emph{International Conference on Learning Representations
  (ICLR)}, 2022.
\newblock arXiv:2111.06236.

\bibitem[Domingos(2025)]{domingos2025tensorlogic}
Pedro Domingos.
\newblock Tensor logic: The language of {AI}.
\newblock \emph{arXiv preprint arXiv:2510.12269}, 2025.
\newblock URL \url{https://arxiv.org/abs/2510.12269}.

\bibitem[Dong et~al.(2021)Dong, Cordonnier, and Loukas]{dong2021attention}
Yihe Dong, Jean-Baptiste Cordonnier, and Andreas Loukas.
\newblock Attention is not all you need: Pure attention loses rank doubly
  exponentially with depth.
\newblock In \emph{Proceedings of the 38th International Conference on Machine
  Learning (ICML)}, pages 2793--2803, 2021.

\bibitem[Dziri et~al.(2023)Dziri, Lu, Sclar, Li, Jiang, Lin, West, Bhagavatula,
  Le~Bras, Hwang, Sanyal, Welleck, Ren, Ettinger, Harchaoui, and
  Choi]{dziri2023faith}
Nouha Dziri, Ximing Lu, Melanie Sclar, Xiang~Lorraine Li, Liwei Jiang,
  Bill~Yuchen Lin, Peter West, Chandra Bhagavatula, Ronan Le~Bras, Jena~D.
  Hwang, Soumya Sanyal, Sean Welleck, Xiang Ren, Allyson Ettinger, Zaid
  Harchaoui, and Yejin Choi.
\newblock Faith and fate: Limits of transformers on compositionality.
\newblock In \emph{Advances in Neural Information Processing Systems 36
  (NeurIPS)}, 2023.

\bibitem[Ellis et~al.(2020)Ellis, Wong, Nye, Sable-Meyer, Cary, Morales,
  Hewitt, Solar-Lezama, and Tenenbaum]{ellis2021dreamcoder}
Kevin Ellis, Catherine Wong, Maxwell Nye, Mathias Sable-Meyer, Luc Cary, Lucas
  Morales, Luke Hewitt, Armando Solar-Lezama, and Joshua~B. Tenenbaum.
\newblock {DreamCoder}: Growing generalizable, interpretable knowledge with
  wake-sleep {B}ayesian program learning.
\newblock \emph{arXiv preprint arXiv:2006.08381}, 2020.
\newblock URL \url{https://arxiv.org/abs/2006.08381}.

\bibitem[Evans and Grefenstette(2018)]{evans2018learning}
Richard Evans and Edward Grefenstette.
\newblock Learning explanatory rules from noisy data.
\newblock \emph{Journal of Artificial Intelligence Research}, 61:\penalty0
  1--64, 2018.

\bibitem[Fodor and Pylyshyn(1988)]{fodor1988connectionism}
Jerry~A. Fodor and Zenon~W. Pylyshyn.
\newblock Connectionism and cognitive architecture: A critical analysis.
\newblock \emph{Cognition}, 28\penalty0 (1-2):\penalty0 3--71, 1988.
\newblock \doi{10.1016/0010-0277(88)90031-5}.

\bibitem[Franz et~al.(2019)Franz, Gogulya, and L{\"o}ffler]{franz2019william}
Arthur Franz, Victoria Gogulya, and Michael L{\"o}ffler.
\newblock {WILLIAM}: A monolithic approach to {AGI}.
\newblock In \emph{International Conference on Artificial General
  Intelligence}, pages 33--43. Springer, 2019.

\bibitem[Ganea et~al.(2018)Ganea, B{\'e}cigneul, and
  Hofmann]{ganea2018hyperbolic}
Octavian-Eugen Ganea, Gary B{\'e}cigneul, and Thomas Hofmann.
\newblock Hyperbolic neural networks.
\newblock In \emph{Advances in Neural Information Processing Systems 31
  (NeurIPS)}, 2018.

\bibitem[Graves(2016)]{graves2016adaptive}
Alex Graves.
\newblock Adaptive computation time for recurrent neural networks, 2016.
\newblock URL \url{https://arxiv.org/abs/1603.08983}.

\bibitem[Graves et~al.(2014)Graves, Wayne, and Danihelka]{graves2014neural}
Alex Graves, Greg Wayne, and Ivo Danihelka.
\newblock Neural turing machines, 2014.
\newblock URL \url{https://arxiv.org/abs/1410.5401}.

\bibitem[Graves et~al.(2016)Graves, Wayne, Reynolds, Harley, Danihelka,
  Grabska-Barwi\'{n}ska, Colmenarejo, Grefenstette, Ramalho, Agapiou, Badia,
  Hermann, Zwols, Ostrovski, Cain, King, Summerfield, Blunsom, Kavukcuoglu, and
  Hassabis]{graves2016hybrid}
Alex Graves, Greg Wayne, Malcolm Reynolds, Tim Harley, Ivo Danihelka, Agnieszka
  Grabska-Barwi\'{n}ska, Sergio~G\'{o}mez Colmenarejo, Edward Grefenstette,
  Tiago Ramalho, John Agapiou, Adri\`{a}~Puigdom\`{e}nech Badia, Karl~Moritz
  Hermann, Yori Zwols, Georg Ostrovski, Adam Cain, Helen King, Christopher
  Summerfield, Phil Blunsom, Koray Kavukcuoglu, and Demis Hassabis.
\newblock Hybrid computing using a neural network with dynamic external memory.
\newblock \emph{Nature}, 538\penalty0 (7626):\penalty0 471--476, 2016.
\newblock \doi{10.1038/nature20101}.

\bibitem[Greff et~al.(2020)Greff, van Steenkiste, and
  Schmidhuber]{greff2020binding}
Klaus Greff, Sjoerd van Steenkiste, and J\"{u}rgen Schmidhuber.
\newblock On the binding problem in artificial neural networks, 2020.
\newblock URL \url{https://arxiv.org/abs/2012.05208}.

\bibitem[Greydanus et~al.(2019)Greydanus, Dzamba, and
  Yosinski]{greydanus2019hamiltonian}
Sam Greydanus, Misko Dzamba, and Jason Yosinski.
\newblock Hamiltonian neural networks.
\newblock In \emph{Advances in Neural Information Processing Systems
  (NeurIPS)}, 2019.
\newblock URL \url{https://arxiv.org/abs/1906.01563}.

\bibitem[Hein et~al.(2019)Hein, Andriushchenko, and Bitterwolf]{hein2019relu}
Matthias Hein, Maksym Andriushchenko, and Julian Bitterwolf.
\newblock Why {ReLU} networks yield high-confidence predictions far away from
  the training data and how to mitigate the problem.
\newblock In \emph{Proceedings of the IEEE/CVF Conference on Computer Vision
  and Pattern Recognition (CVPR)}, pages 41--50, 2019.
\newblock \doi{10.1109/CVPR.2019.00013}.

\bibitem[Hornik et~al.(1989)Hornik, Stinchcombe, and
  White]{hornik1989multilayer}
Kurt Hornik, Maxwell Stinchcombe, and Halbert White.
\newblock Multilayer feedforward networks are universal approximators.
\newblock \emph{Neural Networks}, 2\penalty0 (5):\penalty0 359--366, 1989.
\newblock \doi{10.1016/0893-6080(89)90020-8}.

\bibitem[Jouppi et~al.(2017)Jouppi, Young, Patil, Patterson,
  et~al.]{jouppi2017tpu}
Norman~P. Jouppi, Cliff Young, Nishant Patil, David Patterson, et~al.
\newblock In-datacenter performance analysis of a tensor processing unit.
\newblock In \emph{Proceedings of the 44th Annual International Symposium on
  Computer Architecture (ISCA)}, 2017.
\newblock \doi{10.1145/3079856.3080246}.

\bibitem[Kang et~al.(2024)Kang, Setlur, Tomlin, and
  Levine]{kangocs2024extrapolate}
Katie Kang, Amrith Setlur, Claire Tomlin, and Sergey Levine.
\newblock Deep neural networks tend to extrapolate predictably, 2024.
\newblock URL \url{https://arxiv.org/abs/2310.00873}.

\bibitem[Kautz(2022)]{kautz2022third}
Henry~A. Kautz.
\newblock The third {AI} summer: {AAAI} {Robert S. Engelmore} memorial lecture.
\newblock \emph{AI Magazine}, 43\penalty0 (1):\penalty0 93--104, 2022.
\newblock \doi{10.1609/aimag.v43i1.19122}.

\bibitem[Kim(2025)]{kim2025standard}
Youngsung Kim.
\newblock Standard neural computation alone is insufficient for logical
  intelligence, 2025.
\newblock URL \url{https://arxiv.org/abs/2502.02135}.

\bibitem[Kipf and Welling(2017)]{kipf2017semisupervised}
Thomas~N. Kipf and Max Welling.
\newblock Semi-supervised classification with graph convolutional networks.
\newblock In \emph{International Conference on Learning Representations
  (ICLR)}, 2017.

\bibitem[Koller and Friedman(2009)]{koller2009probabilistic}
Daphne Koller and Nir Friedman.
\newblock \emph{Probabilistic Graphical Models: Principles and Techniques}.
\newblock MIT Press, Cambridge, MA, 2009.
\newblock ISBN 9780262013192.

\bibitem[Kowalski(1979)]{kowalski1979algorithm}
Robert Kowalski.
\newblock Algorithm = logic + control.
\newblock \emph{Communications of the ACM}, 22\penalty0 (7):\penalty0 424--436,
  1979.

\bibitem[Koza(1992)]{koza1992genetic}
John~R. Koza.
\newblock \emph{Genetic Programming: On the Programming of Computers by Means
  of Natural Selection}.
\newblock MIT Press, 1992.

\bibitem[La~Cava et~al.(2021)La~Cava, Orzechowski, Burlacu, de~Fran{\c{c}}a,
  Virgolin, Jin, Kommenda, and Moore]{lacava2021srbench}
William La~Cava, Patryk Orzechowski, Bogdan Burlacu, Fabr{\'i}cio~Olivetti
  de~Fran{\c{c}}a, Marco Virgolin, Ying Jin, Michael Kommenda, and Jason~H.
  Moore.
\newblock Contemporary symbolic regression methods and their relative
  performance.
\newblock In \emph{Advances in Neural Information Processing Systems, Datasets
  and Benchmarks Track}, 2021.

\bibitem[Lakshminarayanan et~al.(2017)Lakshminarayanan, Pritzel, and
  Blundell]{lakshminarayanan2017ensembles}
Balaji Lakshminarayanan, Alexander Pritzel, and Charles Blundell.
\newblock Simple and scalable predictive uncertainty estimation using deep
  ensembles.
\newblock In \emph{Advances in Neural Information Processing Systems
  (NeurIPS)}, 2017.
\newblock URL \url{https://arxiv.org/abs/1612.01474}.

\bibitem[Law et~al.(2014)Law, Russo, and Broda]{law2014answersetprograms}
Mark Law, Alessandra Russo, and Krysia Broda.
\newblock Inductive learning of answer set programs.
\newblock In \emph{European Conference on Logics in Artificial Intelligence
  (JELIA)}, pages 311--325. Springer, 2014.

\bibitem[Lee et~al.(2018)Lee, Lee, Lee, and Shin]{lee2018simple}
Kimin Lee, Kibok Lee, Honglak Lee, and Jinwoo Shin.
\newblock A simple unified framework for detecting out-of-distribution samples
  and adversarial attacks.
\newblock In \emph{Advances in Neural Information Processing Systems
  (NeurIPS)}, 2018.
\newblock URL \url{https://arxiv.org/abs/1807.03888}.

\bibitem[Lee et~al.(2025)Lee, Sim, Shin, Seo, Park, Lee, Hwang, Kim, and
  Kim]{lee2025loth}
Seungpil Lee, Woochang Sim, Donghyeon Shin, Wongyu Seo, Jiwon Park, Seokki Lee,
  Sanha Hwang, Sejin Kim, and Sundong Kim.
\newblock Reasoning abilities of large language models: In-depth analysis on
  the {Abstraction} and {Reasoning} {Corpus}.
\newblock \emph{ACM Transactions on Intelligent Systems and Technology}, 2025.
\newblock \doi{10.1145/3712701}.

\bibitem[Li et~al.(2024)Li, Hu, Larsen, Wu, Alford, Woo, Dunn, Tang, Naim,
  Nguyen, Zheng, Tavares, Pu, and Ellis]{li2024induction}
Wen-Ding Li, Keya Hu, Carter Larsen, Yuqing Wu, Simon Alford, Caleb Woo,
  Spencer~M. Dunn, Hao Tang, Michelangelo Naim, Dat Nguyen, Wei-Long Zheng,
  Zenna Tavares, Yewen Pu, and Kevin Ellis.
\newblock Combining induction and transduction for abstract reasoning.
\newblock \emph{arXiv preprint arXiv:2411.02272}, 2024.
\newblock URL \url{https://arxiv.org/abs/2411.02272}.

\bibitem[Manhaeve et~al.(2018)Manhaeve, Duman{\v{c}}i{\'c}, Kimmig, Demeester,
  and De~Raedt]{manhaeve2018deepproblog}
Robin Manhaeve, Sebastijan Duman{\v{c}}i{\'c}, Angelika Kimmig, Thomas
  Demeester, and Luc De~Raedt.
\newblock {DeepProbLog}: Neural probabilistic logic programming.
\newblock In \emph{Advances in Neural Information Processing Systems 31
  (NeurIPS)}, 2018.

\bibitem[Marcus(2018)]{marcus2018deep}
Gary Marcus.
\newblock Deep learning: A critical appraisal, 2018.
\newblock URL \url{https://arxiv.org/abs/1801.00631}.

\bibitem[Marcus(1998)]{marcus1998rethinking}
Gary~F. Marcus.
\newblock Rethinking eliminative connectionism.
\newblock \emph{Cognitive Psychology}, 37\penalty0 (3):\penalty0 243--282,
  1998.

\bibitem[Marcus(2001)]{marcus2001algebraic}
Gary~F. Marcus.
\newblock \emph{The Algebraic Mind: Integrating Connectionism and Cognitive
  Science}.
\newblock MIT Press, 2001.

\bibitem[McCarthy(1988)]{mccarthy1988epistemological}
John McCarthy.
\newblock Epistemological challenges for connectionism.
\newblock \emph{Behavioral and Brain Sciences}, 11\penalty0 (1):\penalty0 44,
  1988.

\bibitem[Mirzadeh et~al.(2024)Mirzadeh, Alizadeh, Shahrokhi, Tuzel, Bengio, and
  Farajtabar]{mirzadeh2024gsmsymbolic}
Iman Mirzadeh, Keivan Alizadeh, Hooman Shahrokhi, Oncel Tuzel, Samy Bengio, and
  Mehrdad Farajtabar.
\newblock {GSM}-symbolic: Understanding the limitations of mathematical
  reasoning in large language models, 2024.
\newblock URL \url{https://arxiv.org/abs/2410.05229}.

\bibitem[Mont\'{u}far et~al.(2014)Mont\'{u}far, Pascanu, Cho, and
  Bengio]{montufar2014number}
Guido~F. Mont\'{u}far, Razvan Pascanu, Kyunghyun Cho, and Yoshua Bengio.
\newblock On the number of linear regions of deep neural networks.
\newblock In \emph{Advances in Neural Information Processing Systems 27
  (NeurIPS)}, pages 2924--2932, 2014.

\bibitem[Muggleton(1991)]{muggleton1991inductive}
Stephen Muggleton.
\newblock Inductive logic programming.
\newblock \emph{New Generation Computing}, 8\penalty0 (4):\penalty0 295--318,
  1991.

\bibitem[Muggleton(1995)]{muggleton1995inverse}
Stephen Muggleton.
\newblock Inverse entailment and progol.
\newblock \emph{New Generation Computing}, 13\penalty0 (3-4):\penalty0
  245--286, 1995.
\newblock \doi{10.1007/BF03037227}.

\bibitem[Muggleton et~al.(2015)Muggleton, Lin, and
  Tamaddoni-Nezhad]{muggleton2015metainterpretive}
Stephen~H. Muggleton, Dianhuan Lin, and Alireza Tamaddoni-Nezhad.
\newblock Meta-interpretive learning of higher-order dyadic datalog: predicate
  invention revisited.
\newblock \emph{Machine Learning}, 100\penalty0 (1):\penalty0 49--73, 2015.

\bibitem[Ovadia et~al.(2019)Ovadia, Fertig, Ren, Nado, Sculley, Nowozin,
  Dillon, Lakshminarayanan, and Snoek]{ovadia2019trust}
Yaniv Ovadia, Emily Fertig, Jie Ren, Zachary Nado, D.~Sculley, Sebastian
  Nowozin, Joshua~V. Dillon, Balaji Lakshminarayanan, and Jasper Snoek.
\newblock Can you trust your model's uncertainty? evaluating predictive
  uncertainty under dataset shift.
\newblock In \emph{Advances in Neural Information Processing Systems
  (NeurIPS)}, 2019.
\newblock URL \url{https://arxiv.org/abs/1906.02530}.

\bibitem[Rocha et~al.(2020{\natexlab{a}})Rocha, Santos~Costa, and
  Reis]{rocha2020overcomingrl}
Filipe~Marinho Rocha, V{\'i}tor Santos~Costa, and Lu{\'i}s~Paulo Reis.
\newblock Overcoming reinforcement learning limits with inductive logic
  programming.
\newblock In \emph{Trends and Innovations in Information Systems and
  Technologies (WorldCIST 2020)}, volume 1160 of \emph{Advances in Intelligent
  Systems and Computing}, pages 414--423. Springer, Cham, 2020{\natexlab{a}}.
\newblock \doi{10.1007/978-3-030-45691-7_38}.

\bibitem[Rocha et~al.(2020{\natexlab{b}})Rocha, Santos~Costa, and
  Reis]{rocha2020towardsagi}
Filipe~Marinho Rocha, V{\'i}tor Santos~Costa, and Lu{\'i}s~Paulo Reis.
\newblock From reinforcement learning towards artificial general intelligence.
\newblock In \emph{Trends and Innovations in Information Systems and
  Technologies (WorldCIST 2020)}, volume 1160 of \emph{Advances in Intelligent
  Systems and Computing}, pages 401--413. Springer, Cham, 2020{\natexlab{b}}.
\newblock \doi{10.1007/978-3-030-45691-7_37}.

\bibitem[Rocha et~al.(2025)Rocha, Dutra, Santos~Costa, and
  Reis]{rocha2024programsynthesis}
Filipe~Marinho Rocha, In{\^e}s Dutra, V{\'i}tor Santos~Costa, and
  Lu{\'i}s~Paulo Reis.
\newblock Program synthesis using inductive logic programming for the
  abstraction and reasoning corpus.
\newblock \emph{Intelligenza Artificiale}, 19:\penalty0 85--101, 2025.
\newblock \doi{10.1177/17248035251363178}.

\bibitem[Rocha et~al.(2026)Rocha, Dutra, Santos~Costa, and
  Reis]{rocha2026certificate}
Filipe~Marinho Rocha, In\^{e}s Dutra, V\'{i}tor Santos~Costa, and
  Lu\'{i}s~Paulo Reis.
\newblock Certifying extrapolation from the training set alone: Leverage
  blocks, depth, and a precision floor, 2026.
\newblock Manuscript in preparation.

\bibitem[Santoro et~al.(2017)Santoro, Raposo, Barrett, Malinowski, Pascanu,
  Battaglia, and Lillicrap]{santoro2017simple}
Adam Santoro, David Raposo, David~G. Barrett, Mateusz Malinowski, Razvan
  Pascanu, Peter Battaglia, and Timothy Lillicrap.
\newblock A simple neural network module for relational reasoning.
\newblock In \emph{Advances in Neural Information Processing Systems 30
  (NeurIPS)}, pages 4967--4976, 2017.

\bibitem[Schlichtkrull et~al.(2018)Schlichtkrull, Kipf, Bloem, van~den Berg,
  Titov, and Welling]{schlichtkrull2018modeling}
Michael Schlichtkrull, Thomas~N. Kipf, Peter Bloem, Rianne van~den Berg, Ivan
  Titov, and Max Welling.
\newblock Modeling relational data with graph convolutional networks.
\newblock In \emph{European Semantic Web Conference (ESWC)}, pages 593--607.
  Springer, 2018.
\newblock \doi{10.1007/978-3-319-93417-4_38}.

\bibitem[Schmidt and Lipson(2009)]{schmidt2009distilling}
Michael Schmidt and Hod Lipson.
\newblock Distilling free-form natural laws from experimental data.
\newblock \emph{Science}, 324\penalty0 (5923):\penalty0 81--85, 2009.
\newblock \doi{10.1126/science.1165893}.

\bibitem[Schwarz(1978)]{schwarz1978bic}
Gideon Schwarz.
\newblock Estimating the dimension of a model.
\newblock \emph{The Annals of Statistics}, 6\penalty0 (2):\penalty0 461--464,
  1978.
\newblock \doi{10.1214/aos/1176344136}.

\bibitem[Sitzmann et~al.(2020)Sitzmann, Martel, Bergman, Lindell, and
  Wetzstein]{sitzmann2020siren}
Vincent Sitzmann, Julien N.~P. Martel, Alexander~W. Bergman, David~B. Lindell,
  and Gordon Wetzstein.
\newblock Implicit neural representations with periodic activation functions.
\newblock In \emph{Advances in Neural Information Processing Systems 33
  (NeurIPS)}, 2020.
\newblock URL \url{https://arxiv.org/abs/2006.09661}.

\bibitem[Smolensky(1990)]{smolensky1990tensor}
Paul Smolensky.
\newblock Tensor product variable binding and the representation of symbolic
  structures in connectionist systems.
\newblock \emph{Artificial Intelligence}, 46\penalty0 (1--2):\penalty0
  159--216, 1990.

\bibitem[Sun et~al.(2022)Sun, Ming, Zhu, and Li]{sun2022knn}
Yiyou Sun, Yifei Ming, Xiaojin Zhu, and Yixuan Li.
\newblock Out-of-distribution detection with deep nearest neighbors.
\newblock In \emph{International Conference on Machine Learning (ICML)}, 2022.
\newblock URL \url{https://arxiv.org/abs/2204.06507}.

\bibitem[Teru et~al.(2020)Teru, Denis, and Hamilton]{teru2020grail}
Komal~K. Teru, Etienne Denis, and William~L. Hamilton.
\newblock Inductive relation prediction by subgraph reasoning.
\newblock In \emph{Proceedings of the 37th International Conference on Machine
  Learning (ICML)}, volume 119 of \emph{PMLR}, pages 9448--9457, 2020.
\newblock arXiv:1911.06962.

\bibitem[Udrescu and Tegmark(2020)]{udrescu2020aifeynman}
Silviu-Marian Udrescu and Max Tegmark.
\newblock {AI Feynman}: A physics-inspired method for symbolic regression.
\newblock \emph{Science Advances}, 6\penalty0 (16):\penalty0 eaay2631, 2020.
\newblock URL \url{https://arxiv.org/abs/1905.11481}.

\bibitem[Vaswani et~al.(2017)Vaswani, Shazeer, Parmar, Uszkoreit, Jones, Gomez,
  Kaiser, and Polosukhin]{vaswani2017attention}
Ashish Vaswani, Noam Shazeer, Niki Parmar, Jakob Uszkoreit, Llion Jones,
  Aidan~N. Gomez, Lukasz Kaiser, and Illia Polosukhin.
\newblock Attention is all you need.
\newblock In \emph{Advances in Neural Information Processing Systems 30
  (NeurIPS)}, pages 5998--6008, 2017.

\bibitem[Villar et~al.(2021)Villar, Hogg, Storey-Fisher, Yao, and
  Blum-Smith]{villar2021scalars}
Soledad Villar, David~W. Hogg, Kate Storey-Fisher, Weichi Yao, and Ben
  Blum-Smith.
\newblock Scalars are universal: Equivariant machine learning, structured like
  classical physics.
\newblock \emph{Advances in Neural Information Processing Systems (NeurIPS)},
  2021.
\newblock URL \url{https://arxiv.org/abs/2106.06610}.

\bibitem[Virgolin and Pissis(2022)]{virgolin2022srnphard}
Marco Virgolin and Solon~P. Pissis.
\newblock Symbolic regression is {NP}-hard.
\newblock \emph{Transactions on Machine Learning Research}, 2022.
\newblock URL \url{https://openreview.net/forum?id=LTiaPxqe2e}.

\bibitem[Webb et~al.(2021)Webb, Sinha, and Cohen]{webb2021esbn}
Taylor~W. Webb, Ishan Sinha, and Jonathan~D. Cohen.
\newblock Emergent symbols through binding in external memory.
\newblock In \emph{International Conference on Learning Representations
  (ICLR)}, 2021.
\newblock URL \url{https://arxiv.org/abs/2012.14601}.
\newblock arXiv:2012.14601.

\bibitem[Webb et~al.(2024)Webb, Frankland, Altabaa, Segert, Krishnamurthy,
  Campbell, Russin, Giallanza, O'Reilly, Lafferty, and
  Cohen]{webb2024relational}
Taylor~W. Webb, Steven~M. Frankland, Awni Altabaa, Simon Segert, Kamesh
  Krishnamurthy, Declan Campbell, Jacob Russin, Tyler Giallanza, Randall
  O'Reilly, John Lafferty, and Jonathan~D. Cohen.
\newblock The relational bottleneck as an inductive bias for efficient
  abstraction.
\newblock \emph{Trends in Cognitive Sciences}, 28\penalty0 (9):\penalty0
  829--843, 2024.
\newblock \doi{10.1016/j.tics.2024.04.001}.
\newblock arXiv:2309.06629.

\bibitem[Xu et~al.(2020)Xu, Li, Zhang, Du, Kawarabayashi, and
  Jegelka]{xu2020how}
Keyulu Xu, Jingling Li, Mozhi Zhang, Simon~S. Du, Ken-ichi Kawarabayashi, and
  Stefanie Jegelka.
\newblock How neural networks extrapolate: From feedforward to graph neural
  networks.
\newblock In \emph{International Conference on Learning Representations
  (ICLR)}, 2020.

\bibitem[Yarotsky(2017)]{yarotsky2017error}
Dmitry Yarotsky.
\newblock Error bounds for approximations with deep {ReLU} networks.
\newblock \emph{Neural Networks}, 94:\penalty0 103--114, 2017.
\newblock \doi{10.1016/j.neunet.2017.07.002}.

\bibitem[Yehudai et~al.(2021)Yehudai, Fetaya, Meirom, Chechik, and
  Maron]{yehudai2021local}
Gilad Yehudai, Ethan Fetaya, Eli Meirom, Gal Chechik, and Haggai Maron.
\newblock From local structures to size generalization in graph neural
  networks.
\newblock In \emph{Proceedings of the 38th International Conference on Machine
  Learning (ICML)}, pages 11975--11986, 2021.

\bibitem[Zhang et~al.(2016)Zhang, Bengio, Hardt, Recht, and
  Vinyals]{zhang2016understanding}
Chiyuan Zhang, Samy Bengio, Moritz Hardt, Benjamin Recht, and Oriol Vinyals.
\newblock Understanding deep learning requires rethinking generalization, 2016.
\newblock URL \url{https://arxiv.org/abs/1611.03530}.

\end{thebibliography}

\end{document}